\pdfoutput=1
\documentclass[11pt]{article}

\usepackage[]{acl}
\usepackage{booktabs}
\usepackage{makecell}
\usepackage{multirow}
\usepackage{times}
\usepackage{amsmath}
\usepackage{amssymb}
\usepackage{dsfont}
\usepackage{footnote}
\usepackage{xcolor}
\usepackage{amssymb}  
\usepackage{pifont}   
\newcommand{\att}{\text{\checkmark}}  
\newcommand{\unatt}{\text{\ding{55}}} 
\usepackage{soul}
\usepackage{footnote}
\usepackage{xcolor}
\usepackage{soul}
\newcommand{\highlight}[2]{%
    \begingroup
    \definecolor{hlcolor}{HTML}{#1}%
    \sethlcolor{hlcolor}%
    \hl{#2}%
    \endgroup
}
\usepackage{circledsteps}
\usepackage{amsmath}
\definecolor{deterministic84}{HTML}{8A60B0}
\definecolor{deterministic57}{HTML}{C7519C}
\definecolor{deterministic21}{HTML}{D63A3A}
\definecolor{local10}{HTML}{FFBF50}
\definecolor{local5}{HTML}{BCBD22}
\definecolor{evenodd}{HTML}{12A2A8}
\definecolor{control}{HTML}{1F83B4}
\definecolor{reverse}{HTML}{c77dff}

\definecolor{local2}{HTML}{7b2cbf}
\definecolor{local3}{HTML}{52b788}
\definecolor{annd}{HTML}{ff8800}
\definecolor{dnna}{HTML}{ffaa00}
\definecolor{nnda}{HTML}{f1db33}
\definecolor{dann}{HTML}{dfab06}
\definecolor{dnan}{HTML}{da6220}
\definecolor{noshuffle}{HTML}{495057}
\definecolor{random}{HTML}{a2d6f9}
\usepackage{latexsym}
\usepackage[utf8]{inputenc}
\usepackage[T1]{fontenc}
\newcommand{\tokenShe}{%
  \highlight{FFC5D0}{She}%
}
\newcommand{\tokenenjoyed}{%
  \highlight{F1CEB0}{ enjoyed}%
}
\newcommand{\tokenthe}{%
  \highlight{D9D7A7}{ the}%
}
\newcommand{\tokenthree}{%
  \highlight{A1E2C7}{ three}%
}
\newcommand{\tokenfant}{%
  \highlight{9AE1E1}{ fant}%
}
\newcommand{\tokenastically}{%
  \highlight{B8D8F8}{astically}%
}
\newcommand{\tokeninteresting}{%
  \highlight{C4D5FB}{ interesting\,}%
}

\newcommand{\tokenbooks}{%
  \highlight{DACEFB}{ books}%
}

\newcommand{\tokena}{%
  \highlight{F4C6EF}{ a}%
}
\newcommand{\tokenalot}{%
  \highlight{FFC5D9}{ lot}%
}
\newcommand{\tokenperiod}{%
  \highlight{C87A8A}{ .}%
}
\usepackage[utf8]{inputenc}

\usepackage{microtype}

\usepackage{inconsolata}

\usepackage{graphicx}

\title{Anything Goes? A Crosslinguistic Study of (Im)possible Language Learning in LMs}

\author{
    Xiulin Yang\textsuperscript{$\alpha$} \quad
    Tatsuya Aoyama\textsuperscript{$\alpha$} \quad
    Yuekun Yao\textsuperscript{$\beta$} \quad
    Ethan Gotlieb Wilcox\textsuperscript{$\alpha$}
    \\[1ex]
    \textsuperscript{$\alpha$}Georgetown University \quad
    \textsuperscript{$\beta$}Saarland University
    \\
    {
        \texttt{\{xy236, ta571, ethan.wilcox\}@georgetown.edu} \quad
        \texttt{ykyao@coli.uni-saarland.de}
    }
}
\begin{document}
\maketitle
\begin{abstract}
Do language models (LMs) offer insights into human language learning? A common argument against this idea is that because their architecture and training paradigm are so vastly different from humans, LMs can learn arbitrary inputs as easily as natural languages. 
We test this claim by training LMs to model impossible and typologically unattested languages.
Unlike previous work, which has focused exclusively on English, we conduct experiments on 12 languages from 4 language families with two newly constructed parallel corpora. Our results show that while GPT-2 small can largely distinguish attested languages from their impossible counterparts, it does not achieve perfect separation between all the attested languages and all the impossible ones. We further test whether GPT-2 small distinguishes typologically attested from unattested languages with different NP orders by manipulating word order based on Greenberg's Universal 20. We find that the model's perplexity scores do not distinguish attested vs. unattested word orders, while its performance on the generalization test does. These findings suggest that LMs exhibit some human-like inductive biases, though these biases are weaker than those found in human learners.

\end{abstract}

\section{Introduction}
To what extent can language models (LMs) serve as models of human language acquisition and processing? Some, such as \citet{piantadosi2023modern}, argue that LMs can function as comprehensive linguistic theories, challenging traditional symbolic generative approaches. However, critics maintain that the success of LMs is largely irrelevant to human cognition due to fundamental differences in architecture and learning mechanisms \citep{chomsky2023false,fox2024large}. Moreover, studies have shown that LMs fail to acquire key aspects of linguistic knowledge, suggesting that they are limited as models of human language \citep{fox2024large, lan2024large, katzir2023large, dentella2024language}. One central argument in this debate is that LMs are highly flexible learners, capable of acquiring linguistic patterns beyond those learnable by humans, thus making the ability of LMs to learn human languages uninformative for understanding human language acquisition \citep{chomsky2022secrets,moro2023embodied,moro2023large}. 

We present data favoring a more moderate stance, in line with other recent contributions \citep{futrell2025linguistics, milliere2024language, pater2019generative}. Specifically, we present new empirical evidence from the study of \textit{impossible languages} \citep{kallini-etal-2024-mission} in a multilingual setting, suggesting that LMs exhibit some learning biases that align with certain aspects of human cognition. At the same time, their learning behavior is not universally human-like, suggesting that they have simultaneous biases (or a lack thereof) that diverge from human language processing.

We focus on LMs' abilities to learn different types of languages, both possible (attested or unattested) and impossible (unattested by definition). Specifically, for possible languages, we define \textbf{attested languages} as the natural languages spoken by humans (e.g., English, German, and Chinese); \textbf{unattested languages} as languages constructed on language universals and identified in typological studies as \textit{never-occurring}. We consider \textbf{impossible languages} as those that humans cannot acquire and would never produce. Following \citet{kallini-etal-2024-mission}, we select impossible variants as uncontroversial examples of linguistic impossibility, such as languages with shuffled or reversed word orders. To explore unattested languages, we draw from Greenberg’s Universal 20 \citep{greenberg1963some}, which identifies unattested word order patterns in noun phrases (e.g., adjective-number-determiner-noun). While there is no direct evidence that such languages are unlearnable, previous studies suggest that typological feature frequencies correlate with learnability in human learners \citep{culbertson2020learning,gentner2009some,saffran2008grammatical}.

Regarding impossible language modeling, \citet{kallini-etal-2024-mission} provided initial evidence that GPT-2 small can distinguish between possible and impossible variants of English, suggesting that transformer models encode human-like linguistic biases \citep{futrell2025linguistics}. However, their study was limited to English, leaving the question of whether this finding generalizes across languages unanswered. Furthermore, their focus on impossible languages leaves the study of unattested languages largely unexplored (although see \citet{xu2025can} for recent work in this area).

This paper is organized around two main research questions: (1) \textbf{Does LMs' learning behavior distinguish between attested and impossible languages?} Specifically, (a) Within each attested language, do LMs demonstrate better learning of an attested language compared to its impossible variants? (b) Across different attested languages from multiple language families, do LMs demonstrate better learning of \emph{all} attested languages compared to \emph{all} impossible languages? (2) \textbf{Does LMs' learning behavior distibguish between attested and unattested languages?} Specifically, does LMs' ability to model unattested languages align with human typological biases? 

Our experiments on two parallel corpora show that GPT-2 is better at language modeling attested compared to impossible languages in most settings, though this distinction weakens for certain locally shuffled variants in some languages (1a). However, the models' learning behavior does not distinguish attested from impossible languages across languages (1b). It assigns lower perplexity to unattested languages with preserved constituency and fixed word order, yet performs better on typologically attested languages in the generalization test (2). These findings suggest that LMs show certain human-like learning biases \citep[e.g.,][]{culbertson2020learning}, though not full alignment.\footnote{Our code and data are available at \url{https://github.com/picol-georgetown/multilingual-LM}.}

\section{Related Work}
\label{literature_review}
\subsection{Language Models \& Cognitive Plausibility}
Recent advances in deep learning have led to an upsurge in cognitive modeling with artificial neural networks, especially for language \citep[e.g.,][]{wilcox-etal-2023-testing,borenstein-etal-2024-languages,kirov-cotterell-2018-recurrent}. However, linguists remain divided on whether LMs can meaningfully inform linguistic theories. 
On the one hand, LMs are limited: They lack the capacity for (compositional) generalization \citep{yao-koller-2022-structural,kim-linzen-2020-cogs} and display biases inconsistent with human learning and processing of certain linguistic phenomena \citep{de-dios-flores-etal-2023-dependency,davis-van-schijndel-2020-recurrent,mitchell-bowers-2020-priorless}. These issues suggest that, beyond functioning as sophisticated probability estimators, LMs have limited use as cognitive models \citep{cuskley2024limitations,bolhuis2024three,chomsky2023false}. Of particular relevance to our study is the argument that LMs can learn patterns that are difficult or even impossible for humans \citep{chomsky2023false, moro2023large}. This suggests that LMs do not share the cognitive constraints inherent to the human brain and may therefore miss patterns to which humans are naturally biased, rendering them uninformative for understanding human cognition.

On the other hand, LMs have advanced psycholinguistics by serving as highly accurate probability estimators and have already been used to test and refine theories such as Surprisal Theory \citep{goodkind-bicknell-2018-predictive,oh-schuler-2023-surprisal,oh-schuler-2023-transformer,kuribayashi-etal-2024-psychometric}, Uniform Information Density \citep{meister-etal-2021-revisiting,tsipidi-etal-2024-surprise}, and other cognitive-linguistic theories and psychometrics \citep{pearl2011far,gibson2019efficiency,kuribayashi2025large}. More recently, \citet{kallini-etal-2024-mission,xu2025can}'s experiments demonstrate that LMs can distinguish between possible and (typologically) impossible languages \citep{chomsky2023false,moro2023large} in studies focusing on English and Japanese. These findings provide some empirical counter-evidence to the above arguments. 
\looseness=-1

\subsection{Multilingual Language Modeling}

Whether languages vary in complexity remains a controversial topic, and linguists have taken different approaches to address this question \citep[e.g.,][]{mcwhorter2001worlds,mcwhorter2011linguistic,newmeyer2021complexity,joseph2012all}. While most generative linguists argue that Universal Grammar requires that all languages be equally complex, others have challenged this notion \citep{gil2008complex}.\footnote{See \citet{newmeyer2021complexity} for a more thorough discussion.}

Initial computational attempts to examine language complexity using LMs were limited to RNN-based architectures \citep{cotterell-etal-2018-languages,mielke-etal-2019-kind,Johnson2021InvestigatingTE} and $n$-grams \citep{koplenig2023languages}. These studies suggest that language complexity correlates with morphological richness and the size of speaker populations. More recently, \citet{arnett-bergen-2025-language} investigated why morphologically rich languages are harder to model. By testing monolingual LMs trained on carefully curated comparative datasets \citep{chang2024goldfishmonolinguallanguagemodels}, they found that morphological features alone could not predict language learnability when training data size was controlled.

While valuable, previous studies often rely on non-parallel corpora, introducing inconsistencies across languages. Even with parallel corpora \citep{mielke-etal-2019-kind}, studies are limited by small datasets and outdated models. Our study addresses these gaps using a larger parallel corpus and modern transformer architectures.

\section{Data and Implementation Details}

\subsection{Parallel Data Construction: OPUS12 and OPUS30}

One challenge in multilingual comparisons is that texts drawn from different sources in different languages will have different amounts of information. To control for this, we construct two sentence-aligned multilingual parallel corpora to ensure that all languages in our dataset match in terms of content. This allows us to isolate the effect of how formal properties of a language might impact its learnability. \looseness=-1

We name the two parallel corpora \textbf{OPUS12} and \textbf{OPUS30}, gathering aligned sentences from five corpora available on OPUS \citep{tiedemann-2012-parallel}: NLLB \citep{schwenk-etal-2021-ccmatrix}, TED2020 \citep{reimers-gurevych-2020-making}, the Bible \citep{christodouloupoulos2015massively}, OpenSubtitles \citep{lison-tiedemann-2016-opensubtitles2016}, and CCAligned \citep{elkishky_ccaligned_2020}. Since overlap among languages decreases as more languages are included, we decided to select a minimum of 10M words in English as a standard for our parallel corpora. 10M words also correspond to the amount of input of children's first 2 to 5 years of development \citep{warstadt-etal-2023-findings}.

OPUS12 is a 12-language multilingual sentence-aligned corpus\footnote{The languages and their typological information are listed in Appendix~\ref{appendix:data-details}.}. There are around 10M words in the case of English. OPUS30 contains 30 languages with a smaller data size: 48K sentences with 0.7M words. While the two datasets share overlapping languages, their sentences do not overlap, making OPUS30 a suitable test set for additional language modeling experiments. 

After deduplicating and removing English sentences from non-English data split using FastText \citep{joulin-etal-2017-bag}, we report the statistics of our corpora in Table~\ref{tab:data}.

\begin{table}
    \centering
    \small
    \begin{tabular}{lrr|rr}
    \toprule
    Data Source & \multicolumn{2}{c|}{OPUS12} & \multicolumn{2}{c}{OPUS30} \\
    \cmidrule(lr){2-3} \cmidrule(lr){4-5}
                & \# Sent & \# Word & \# Sent & \# Word \\
    \midrule
    NLLB        & 5K      & 0.1M    & 16       & 368       \\
    TED2020     & 164K    & 2.9M    & 11K       & 182K      \\
    Bible       & 40K     & 1M      & 14K      & 324K       \\
    OpenSubtitles & 680K  & 4.5M    & 15K       & 60K      \\
    CCAligned   & 117K    & 1.6M    & 8K       & 111K      \\
    Overall     & 1M      & 10.1M   &  48K     & 0.7M       \\
    \bottomrule
    \end{tabular}
    \caption{Data sources of OPUS12 and OPUS30. The word counts are based on the English data.}
    \label{tab:data}
\end{table}

\subsection{Validation Experiment}
To ensure the reliability of our findings presented in the remainder of this paper, we replicate experiments in \citet{kallini-etal-2024-mission} using a scaled-down version of their original corpus (10M words). We find a perfect rank correlation between our results and theirs ($\text{Spearman’s }\rho=\textbf{1}, p < 0.001$). More information can be found in Appendix~\ref{replication}. 

\subsection{Model Architecture \& Training}

\begin{table*}[h]
\small
    \centering
    \renewcommand{\arraystretch}{1.3} % Adjust row height
    \setlength{\tabcolsep}{8pt} % Adjust column spacing
    \resizebox{\textwidth}{!}{\begin{tabular}{c|l|l}
        \toprule
        \textbf{Group} & \textbf{Language} & \textbf{Definition} \\
        \midrule
        \multirow{2}{*}{\textbf{Ours}} 
            & \textcolor{local2}{\textsc{shuffle\_local (w=2)}} & The sentence is reordered with every two tokens reversed in order. \\
        \cline{2-3}
            & \textcolor{reverse}{\textsc{reverse\_full}} & Every word is reversed in order in a sentence. \\
        \hline
        \multirow{7}{*}{\textbf{\textsc{k}+}} 
            & \textcolor{deterministic84}{\textsc{shuffle\_deterministic (s=84)}} & The sentence is deterministically shuffled by length with seed 84. \\
        \cline{2-3}
            & \textcolor{deterministic57}{\textsc{shuffle\_deterministic (s=57)}} & The sentence is deterministically shuffled by length with seed 57.\\
        \cline{2-3}
            & \textcolor{deterministic21}{\textsc{shuffle\_deterministic (s=21)}} & The sentence is deterministically shuffled by length with seed 21.\\
        \cline{2-3}
            & \textcolor{local10}{\textsc{shuffle\_local (w=10)}} & The sentence is deterministically shuffled in local window size being 10.\\
        \cline{2-3}
            & \textcolor{local5}{\textsc{shuffle\_local (w=5)}} & The sentence is deterministically shuffled in local window size being 5.\\
        \cline{2-3}
            & \textcolor{local3}{\textsc{shuffle\_local (w=3)}} & The sentence is deterministically shuffled in local window size being 3.\\
        \cline{2-3}
            & \textcolor{evenodd}{\textsc{shuffle\_even\_odd}} & The sentence is reordered with even-indexed tokens first, then odd-indexed.\\
        \bottomrule
    \end{tabular}}
    \caption{Overview of impossible languages in our Experiment1 and Experiment2. \textbf{\textsc{k+}} languages are borrowed from \citet{kallini-etal-2024-mission} and the rest are new variants introduced in our experiments.}
    \label{tab:impossible_languages}
\end{table*}

In our experiments, following \citet{kallini-etal-2024-mission}, we trained standard GPT-2 small models for each language and evaluated its performance based on the geometric mean perplexity over a parallel test split of 10K randomly sampled sentences. Due to limited computational resources, we trained each model using 3 random seeds instead of the 5 used in the original study, reduced the maximum training steps from 2000 to 1200 to avoid overfitting, and adjusted the warmup steps proportionally to 120.\footnote{We did not experiment with alternative warmup steps, as \citet{kallini-etal-2024-mission} demonstrated that changing the warmup schedule does not affect the ranking of perplexities for impossible LMs.}

\subsection{Multilingual Tokenization}
Given our multilingual experiments, tokenization is crucial for fair comparison. To avoid bias toward Latin-script languages, which are overrepresented in our study, we opted against using a multilingual tokenizer with a shared vocabulary.

Previous monolingual experiments either set the vocabulary size of tokenizers to be the same across languages \cite{arnett-bergen-2025-language} or applied the formula $0.4 \times |V|$ \cite{koplenig2023large, mielke-etal-2019-kind}, where $|V|$ represents the number of unique word types. We conducted a series of pilot experiments on tokenization and found the latter approach unsuitable for our experimental design. Specifically, the large $|V|$ in morphologically rich languages makes it impractical to train a small model with such a large vocabulary size. Details can be found in Appendix~\ref{pilot_study}. 

Given these considerations, we opted to use pretrained tokenizers. The rationale behind this choice is that when the tokenizer training data is sufficiently large and diverse, the resulting tokenization scheme should be equally good across languages, as long as the tokenizer algorithm and hyperparameters (e.g., vocabulary size, subword strategy) remain the same.\footnote{Although tokenization quality, measured by metrics like compression \citep{schmidt-etal-2024-tokenization} and Rényi entropy \citep{zouhar-etal-2023-tokenization}, has been linked to language modeling performance \citep[e.g.,][]{liang-etal-2023-xlm,goldman-etal-2024-unpacking}, recent studies challenge this connection \citep{arnett-bergen-2025-language}.} While it is difficult to say how \textit{sufficiently large and diverse} a tokenizer training set should be for fair comparison, we consider the size of the training data for GPT-2 \citep{radford2019language} as a reference point, as English was a high-resource language even in 2019 when the paper was published. We believe that this data size is sufficient to minimize differences that tokenization will make across languages. \looseness=-1

One potential concern is that the BPE algorithm might not be optimized for agglutinative languages such as Turkish. However, much literature on cross-linguistic LM comparison adopts BPE tokenizers \citep[e.g.,][]{mielke-etal-2019-kind,arnett-bergen-2025-language}. As an additional check, we use token counts per word (TCW; reported in Appendix ~\ref{tcw} Table~\ref{tab:ctc}) to measure the morphological complexity of a language and report the correlation between TCW and our test-set perplexity. The results show the correlation is not significant (see Section\ref{exp2}), suggesting that the morphological complexity of a language does not substantially impact its learnability in our experiments. \looseness=-1

When selecting pretrained tokenizers, we use \textbf{monolingual BPE} tokenizers,\footnote{However, for Chinese, we follow previous studies \citep{mielke-etal-2019-kind} and use the Chinese-BERT tokenizer.} targeting a vocabulary size of approximately 50k, with exceptions for Romanian, Arabic, and Chinese due to limited model availability. The training data for all other languages is at least as large as the English corpus. The tokenizer details can be found in Appendix~\ref{tokenizers}.

% Specifically, we retrain GPT-2 small models on all the impossible languages they proposed, including \textit{*shuffled} languages, \textit{*hop} languages, and \textit{*reversed} languages. The \textit{*shuffled} languages can be found in Table~\ref{}, and other two types of languages can be found in the appendix. 

 % It is worth noting that since our experiment uses only 10\% of the data compared to \citet{kallini-etal-2024-mission}, the model tends to overfit more easily as training steps increase when trained with the same number of steps.

 % To conclude, in this experiment, we show that 10M words are sufficient enough to replicate the language modeling experiments using 100M words conducted by \citet{kallini-etal-2024-mission}.

\begin{figure*}[t]
    \centering
    \includegraphics[width=\linewidth]{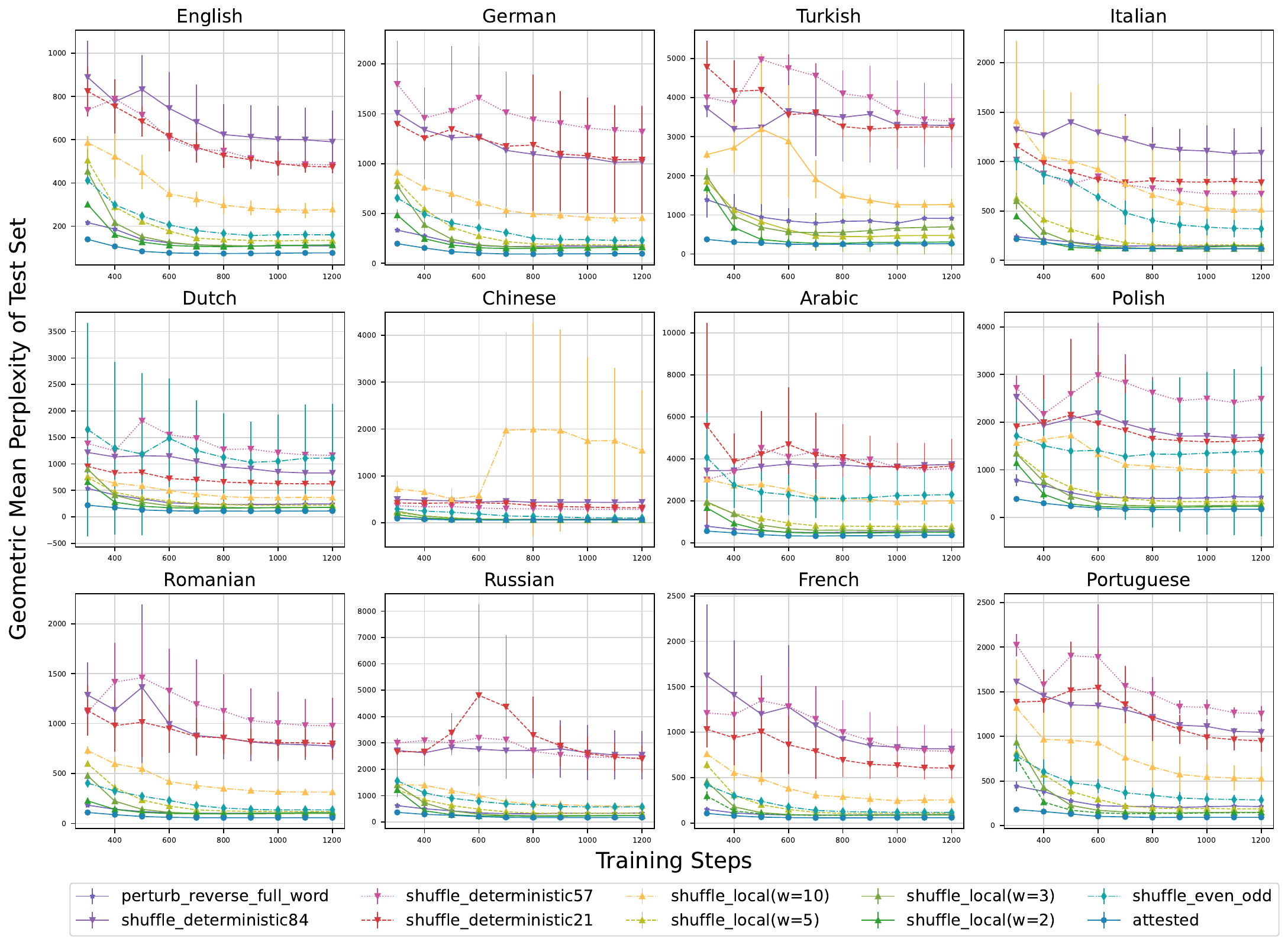}
    \caption{Attested individual Language vs. their corresponding counterparts with a 95\% confidence interval over 3 random seeds tested on 10k sentences from OPUS30.}
    \label{fig:single_language}
\end{figure*}

\section{Experiment 1: Attested vs. Impossible Languages (Intra-Language)}
\subsection{Impossible Languages}  
In this experiment, we use the deterministic shuffled languages from \citet{kallini-etal-2024-mission} along with two new variants (see Table~\ref{tab:impossible_languages}). We include shuffled languages because (1) \citet{kallini-etal-2024-mission} identify them as the \textit{most} impossible languages in their language possibility ranking, and (2) their difficulty is also indirectly supported by empirical studies showing that both adults and children exhibit a regularization bias, which can be thought of as a bias \emph{against} shuffling \citep{newmeyer2005possible,singleton2004learners}.  

Since all languages are deterministically shuffled, the original ones (i.e., attested ones) can be recovered from their variants through another deterministic function. If LMs function as non-human-like pattern recognizers as \citet{chomsky2023false, moro2023large} argue, they should be able to learn these languages as well as attested ones.  

\subsection{Results \& Discussion}

The results of this experiment are presented in Figure~\ref{fig:single_language}.
We note three high-level trends:
First, in all languages except Italian, at every checkpoint, the attested language has a lower mean perplexity than all its impossible variants. For Italian, \textcolor{local2}{\textsc{shuffle\_local (w=2)}} yields a slightly lower perplexity than natural Italian, though the difference is not significant (Mann-Whitney U test: \textit{W} = 63, \textit{p} = 0.353). Welch's t-test with Bonferroni correction across 12 checkpoints shows that for all languages, \textsc{shuffle\_control} differs significantly from other perturbations early in training, but this difference diminishes or becomes insignificant for some languages, especially French, Italian, and Portuguese.\footnote{Details in Appendix~\ref{stats_test}.} Attested languages also show smaller error bars, suggesting more stable learning.

Second, smaller shuffling windows consistently yield lower perplexity. Moreover, \textsc{shuffle\_deterministic} languages result in higher perplexity than \textsc{shuffle\_local}, likely because they shuffle based on sequence length, which autoregressive models cannot directly access. Third, as a sanity check, a Spearman's rank correlation between OPUS30 English and \citet{kallini-etal-2024-mission}'s results shows strong alignment (see Appendix~\ref{replication}).

Based on these findings, we answer the first subquestion: LMs can largely distinguish each attested language from its impossible counterparts by their learning trajectories. 

\section{Experiment 2: Attested vs. Impossible Languages (Inter-Language)}
\label{exp2}
In this experiment, we pool the results of all possible and impossible languages and investigate whether there is a separation boundary between them. If GPT-2 small can distinguish between possible and impossible languages, we expect its perplexity on the former to be lower than on the latter.

\begin{figure}[t]
    \centering
  \includegraphics[width=\linewidth]{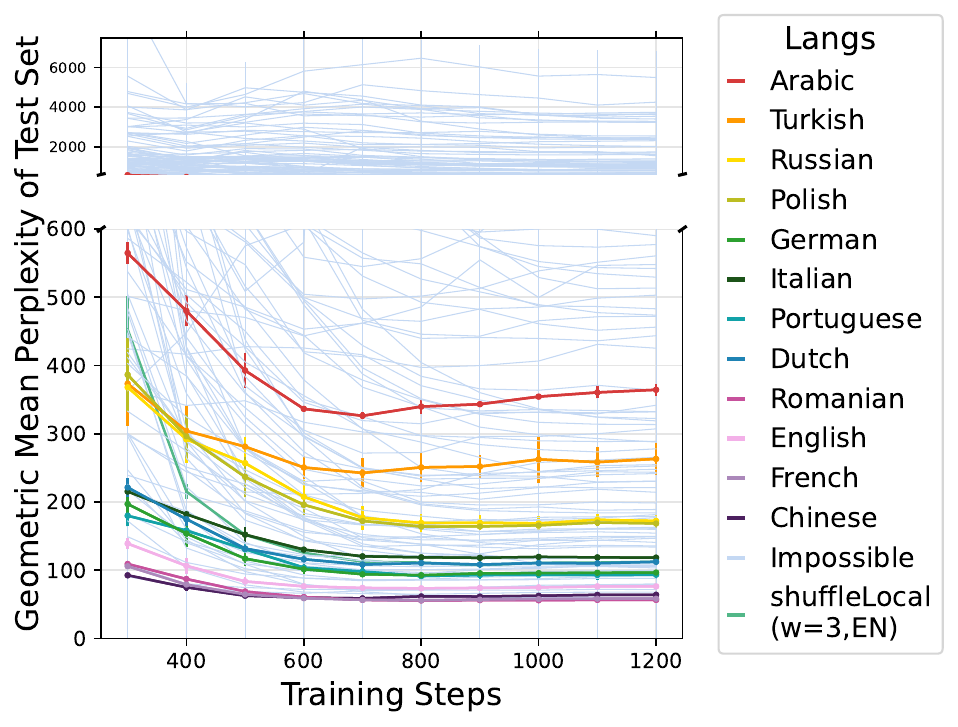}
   \caption{Attested natural languages vs. impossible languages with a 95\% confidence interval over 3 random seeds. The x-axis represents the training steps, and the y-axis shows the perplexity on the test split. All the impossible languages are marked in light blue.}
    \label{fig:multilingual}
\end{figure}

The results of different LMs are shown in Figure~\ref{fig:multilingual}.
%\footnote{To highlight the overlap of perplexity between attested languages and impossible ones, we zoom in on the lower perplexity range while displaying higher perplexity values in a separate, compressed section with a break in the $y$-axis.}
The first thing to note is that not every language shows the same perplexity, with Arabic highest and Chinese the lowest. 

\begin{table*}[ht]
\small
\centering
\begin{tabular}{lllll}
\toprule
 \textbf{Langs} & \multicolumn{2}{c}{\textbf{Attested}} & \textbf{Example}\\
\cmidrule(lr){2-3}
& \textbf{Typo.} & \textbf{Theo.} & \\
\midrule
 \textcolor{nnda}{\textsc{perturb\_\textbf{N}nda}} & \textsc{no} & \textsc{no} & {\texttt{\tokenShe \tokenenjoyed \tokenbooks \tokenthree \tokenthe \tokenfant \tokenastically \tokeninteresting\tokena \tokenalot \tokenperiod}} \\
\textcolor{annd}{\textsc{perturb\_an\textbf{N}d}} & \textsc{no} & \textsc{no} &  \texttt{\tokenShe \tokenenjoyed  \tokenfant \tokenastically \tokeninteresting \tokenthree \tokenbooks \tokenthe \tokena \tokenalot \tokenperiod} \\
 \textcolor{dann}{\textsc{perturb\_da\textbf{N}n}} & \textsc{few} & \textsc{yes} & \texttt{\tokenShe \tokenenjoyed  \tokenthe \tokenfant \tokenastically \tokeninteresting \tokenbooks \tokenthree \tokena \tokenalot \tokenperiod} \\
\textcolor{dnan}{\textsc{dperturb\_dna\textbf{N}}} & \textsc{many} & \textsc{yes} & \texttt{\tokenShe \tokenenjoyed \tokenthe \tokenthree \tokenfant \tokenastically \tokeninteresting \tokenbooks \tokena \tokenalot \tokenperiod} \\
 \textcolor{dnna}{\textsc{perturb\_dn\textbf{N}a}} & \textsc{many} & \textsc{yes} & \texttt{\tokenShe \tokenenjoyed  \tokenthe \tokenthree \tokenbooks \tokenfant \tokenastically \tokeninteresting \tokena   \tokenalot \tokenperiod}\\
\textcolor{random}{\textsc{np\_random}} & \textsc{no} & \textsc{no} &  \texttt{\tokenShe \tokenenjoyed  \tokenbooks  \tokenfant \tokenastically \tokenthree \tokeninteresting  \tokenthe \tokena   \tokenalot \tokenperiod}\\
\bottomrule
\end{tabular}
\caption{\small{List of NP-perturbations with corresponding categories and examples. \textit{Typo} refers to \textit{typologically}-attested, while \textit{Theo} refers to \textit{theoretically}-attested by \citet{cinque2005deriving}'s analysis.}}
\label{tab:new_impossible_languages}
\end{table*} 
We observe a moderate positive correlation between the average number of tokens per word (TCW) and perplexity of each of the last checkpoints in 11 languages (Chinese is excluded because the BERT tokenizer is a character-level tokenizer), as indicated by a Spearman’s rank test ($\rho = 0.564$), but it is not significant ($p = 0.076$). This finding aligns with the observation by \citet{arnett-bergen-2025-language} that there is no significant difference in language modeling difficulty of agglutinative vs. fusional languages when the amount of information is controlled. 

Turning to our main research question, although all the attested languages are distributed at the bottom of the graph, we see that some impossible languages fall between these attested languages. For example, Russian, Turkish, and Arabic all show higher perplexity than English perturbed with \textcolor{local3}{\textsc{shuffle\_local (w=3)}}.
To quantify the extent GPT-2's perplexity values can separate attested from impossible languages, we train a linear SVM classifier with the perplexity value across the three random seeds of each checkpoint as features. The classifier reaches $0.75$ ($sd=0.08$) macro F1 score averaged over 10-folds cross-validation. 

Based on this experiment, we answer the second sub-question posed in our paper: Although LMs tend to learn attested languages better than impossible ones, their perplexity does not distinguish all attested languages from all impossible languages.
%LMs struggle (if not fail) to distinguish all possible languages from all impossible languages overall.

\begin{figure*}
    \centering
    \includegraphics[width=\linewidth]{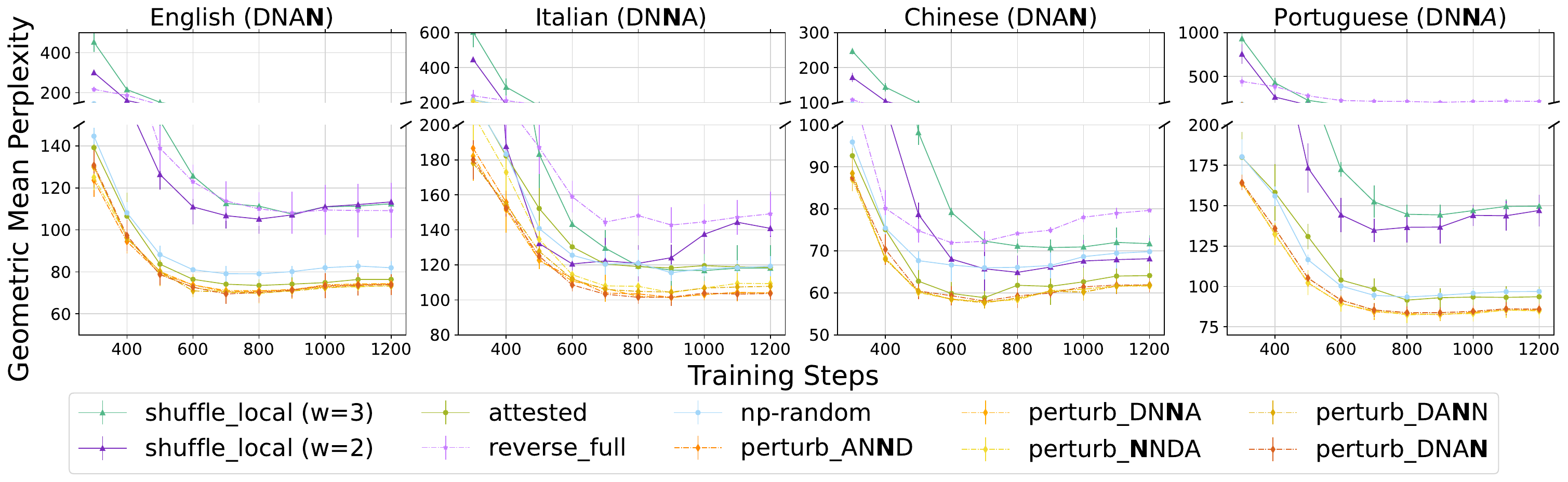}
    \caption{Attested natural languages vs. their corresponding unattested languages with a 95\% confidence interval over 3 random
seeds. The x-axis represents the training steps, and the
y-axis shows the perplexity on the test split. Different language types are distinguished using distinct color palettes. }
    \label{fig:cc}
\end{figure*}

\section{Experiment 3: Attested vs. Unattested Languages}

In this experiment, we investigate how well LMs can learn and generalize to \textbf{unattested languages}, languages whose structure is conceivable according to rules of grammar or morphology, but which have not been found to exist. While unattested languages are not necessarily unlearnable \citep[e.g.,][]{smith1995mind}, prior research suggests a link between typological feature frequency, cognitive biases, and language learnability \citep[e.g.,][]{gentner2009some,culbertson2012learning, culbertson2015harmonic, culbertson2020learning}. 

We focus on Greenberg's Universal 20 \citep{greenberg1963some}, which suggests that certain determiner-adjective-number-noun orders in an NP are universally unattested. \citet{culbertson2015harmonic, culbertson2017innovation, culbertson2020learning} find that harmonic NP orders (i.e., ones where the dependents always all either precede or follow the head; e.g., \textsc{num-adj-noun} and \textsc{noun-adj-num}) are easier to learn than non-harmonic ones (e.g., \textsc{num-noun-adj} or \textsc{adj-noun-num}) for humans. One influential hypothesis, the Typological Prevalence Hypothesis, proposes that more common typological patterns are easier to learn \citep{gentner2009some}. Therefore, if LMs exhibit similar biases as humans, a gradient of difficulty is expected in learning different NP orders, with some unattested configurations posing greater challenges than others.

Among the 24 theoretically possible orders of adjectives, nouns, determiners, and numbers, we select five combinations, covering cases classified as \textsc{few}, \textsc{many}, and \textsc{zero} in \citet{cinque2005deriving}’s typological analysis.\footnote{Although \citet{cinque2005deriving} seeks to explain why \textsc{zero} languages really are ``underivable'' under the minimalist program we refer to them as \emph{unattested} to contrast them with the impossible languages of the previous section, i.e., ones that involve shuffling or reversed word order.} In this experiment, we only permute words within NPs. If the perplexity of these permuted languages is similar to that of attested languages, it suggests two possible reasons: (1) LMs can learn these unattested languages; (2) NPs may be small (in terms of number of tokens) with respect to the entire data size, and hence NP-internal perturbation introduces less noise compared to the entire data perturbation of the previous experiments. 
To rule out the latter possibility, we also construct a control condition in which words corresponding to these POS categories are randomly shuffled within NPs. This language serves as a baseline, indicating the extent to which NP-internal permutations influence the learnability of a language.

Examples of perturbed NP word orders and their typological information are listed in Table~\ref{tab:new_impossible_languages} and their word orders are reported below:

\begin{itemize} \setlength{\itemsep}{0pt} \setlength{\parskip}{0pt}
   \item \textcolor{nnda}{\textsc{perturb\_\textbf{N}nda}}: \textsc{noun>num>det>adj}.
   \item \textcolor{annd}{\textsc{perturb\_an\textbf{N}d}}: \textsc{adj>num>noun>det}.
   \item \textcolor{dann}{\textsc{perturb\_da\textbf{N}n}}: \textsc{det>adj>noun>num}.
   \item \textcolor{dnan}{\textsc{perturb\_dna\textbf{N}}}: \textsc{det>num>adj>noun}, typical of English and Chinese.
   \item \textcolor{dnna}{\textsc{perturb\_dn\textbf{N}a}}: \textsc{det>num>noun>adj}, typical of Italian and Portuguese.
   \item \textcolor{random}{\textsc{np\_random}}: Random permutation of \textsc{adj}, \textsc{noun}, \textsc{num}, and \textsc{det} within NPs.
\end{itemize}

% \begin{figure}
%     \centering
%     \includegraphics[width=\linewidth]{latex/sanity_check.pdf}
% \caption{Language modeling of different impossible languages. The colors of the experiments correspond to the classification shown in Figure~\ref{fig:examples_impossible}. The attested English is highlighted in gray.}
%     \label{fig:test_hypo}
% \end{figure}

Since identifying NP structures requires a constituency parser, we use Stanza \citep{qi-etal-2020-stanza} to parse raw text. Stanza provides constituency parsing for only Chinese, Portuguese, English, and Italian, with acceptable accuracy (>0.85)\footnote{\url{https://stanfordnlp.github.io/stanza/constituency.html}}, so we limit our analysis to these four languages. As different parsers are trained on distinct treebanks with varying annotation guidelines, we select POS tags based on each treebank’s guidelines. Details are provided in Appendix~\ref{treebank}.

\begin{figure*}
    \centering
\includegraphics[width=1.1\linewidth]{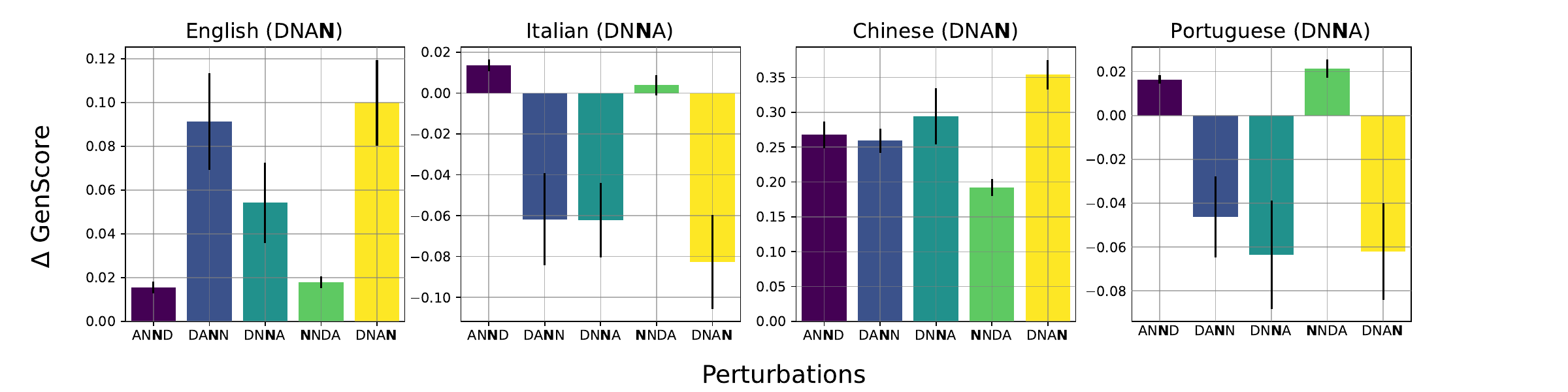}
    \caption{Mean $\Delta$ GenScore across four languages under five NP perturbations. Error bars indicate the 95\% CI computed over three random seeds. Positive $\Delta$ GenScore indicates better generalization for models trained on attested languages, while negative values indicate better generalization for models trained on unattested languages.}
    \label{fig:generalization_test}
\end{figure*}

Studies such as \citet{xu2025can} suggest a difference between models' perplexity and results of targeted evaluations. Therefore, we additionally conduct a targeted test to assess how well LMs trained on different perturbed languages generalize. Specifically, we propose $\Delta$GenScore to quantify their generalization ability, measured across a test corpus of $n$ sentences, and defined as:
\begin{align}
    \Delta \text{GenScore} 
    &= \text{GenScore}_{\att} - \text{GenScore}_{\unatt} \label{eq:delta} \\
    \text{GenScore}_{\att} 
    &= \frac{1}{n} \sum_{i=1}^{n} 
        \mathds{1}{\bigl\{ 
            P_{\att}(s_{\att, i}) > P_{\att}(s_{\unatt, i}) 
        \bigr\}} \nonumber \\
         \text{GenScore}_{\unatt} 
    &= \frac{1}{n} \sum_{i=1}^{n} 
        \mathds{1}{\bigl\{ 
            P_{\unatt}(s_{\unatt, i}) > P_{\unatt}(s_{\att, i}) 
        \bigr\}} \label{eq:unattested} \nonumber
\end{align}

where $\text{GenScore}_{\att}$ refers to the generalization score of a model trained on \textbf{attested (natural) languages}, while $\text{GenScore}_{\unatt}$ refers to the generalization score of a model trained on \textbf{unattested (perturbed) languages}. More specifically, for each test case, we form a minimal pair consisting of an original version $s_{\att, i}$ and its perturbed sentence $s_{\unatt, i}$. Let $P_{\att}$ denote the probability assigned by a model trained on attested languages and $P_{\unatt}$ the probability assigned by a model trained on unattested languages. Then, $\text{GenScore}_{\att}$ is the proportion of cases where $P_{\att}(s_{\att, i}) > P_{\att}(s_{\unatt, i})$, while $\text{GenScore}_{\unatt}$ is the proportion where $P_{\unatt}(s_{\unatt, i}) > P_{\unatt}(s_{\att, i})$. We extract attested sentences from the same test set used for perplexity evaluation but include only those with at least one perturbed NP. The minimal pair test is conducted using the last checkpoint of each language model. 

If a model assigns higher probability to natural (i.e., attested) word orders \emph{regardless of its training data}, then it would obtain a $\Delta$\text{GenScore} of $1$.
Likewise, if it assigns a higher probability to unattested orders regardless of its training data, then it would have a $\Delta$\text{GenScore} of $-1$.
A $\Delta$\text{GenScore} of $0$ indicates that the model always assigns higher probabilities to sequences that match its training data.
Therefore, we interpret positive $\Delta$\text{GenScore} values as indicating better generalization for natural word orderings, and negative scores as indicating better generalization for perturbed orderings.
We use $\Delta$\text{GenScore} to investigate models trained on each of our natural languages, and compare them to each of our possible NP perturbations.

%If a model learns its training data well, it should assign higher probabilities to sentences following the NP patterns seen during training. Thus, models trained on unattested languages should favor sequences containing the corresponding perturbed NPs, and similarly for attested languages. Since perplexity reflects how well a model predicts test data, we expect $\Delta$\text{GenScore} to be positive when attested LMs have lower perplexity than unattested ones (indicating better fit) and negative when unattested LMs have lower perplexity.
\subsection{Results}  
\paragraph{Perplexity} Our results (Figure~\ref{fig:cc}, bottom subgraph) show that shuffling POS tags within NPs increases perplexity, often matching or exceeding that of attested languages. This rules out the possibility that limited perturbations do not affect model training. Surprisingly, all five NP-perturbed languages exhibit lower perplexity than their attested counterparts across all four languages, though the differences are not significant for Italian, Chinese, and Portuguese (by a Welch's t-test with Bonferroni correction).\footnote{For English, there is no significant difference between \textsc{shuffle\_control} and \textcolor{dnan}{\textsc{dna\textbf{N}}}, the dominant NP order in English.} No significant difference is observed between languages with attested (i.e., \textcolor{dann}{\textsc{da\textbf{N}n}}, \textcolor{dnan}{\textsc{dna\textbf{N}}}, and \textcolor{dnna}{\textsc{dn\textbf{N}a}}) and unattested NP orders (i.e., \textcolor{nnda}{\textsc{\textbf{N}nda}} and \textcolor{annd}{\textsc{an\textbf{N}d}}) either, indicating a lack of human alignment in language learning bias.  

\paragraph{Generalization Test} 
The results from this experiment are visualized in Figure~\ref{fig:generalization_test} and present a mixed picture.
%Since unattested English languages have significantly lower perplexity than attested English, we expect a negative $\Delta$GenScore, while for other languages, we expect a slightly negative $\Delta$GenScore due to non-significant but lower perplexity differences. However, Figure~\ref{fig:generalization_test} shows mixed results, consistent with \citet{xu2025can}'s findings of discrepancies between perplexity and targeted evaluation. 
Two observations emerge. First, models trained on \textcolor{nnda}{\textsc{Nnda}} and \textcolor{annd}{\textsc{anNd}}, the two typologically absent orderings, consistently yield positive $\Delta$GenScore across all languages. This indicates poorer generalization of models trained on unattested patterns than models on attested ones listed in Table~\ref{tab:new_impossible_languages}. 
Second, $\Delta$GenScore remains positive for all five NP perturbations in English and Chinese but shows mixed results for Italian and Portuguese. Since English and Chinese predominantly follow the \textcolor{dnan}{\textsc{dnaN}} order and Italian and Portuguese follow \textcolor{dnna}{\textsc{dnNa}}, this suggests models trained on \textcolor{dnan}{\textsc{dnaN}} orders generalize more consistently. This finding, if confirmed, supports \citet{culbertson2015harmonic}'s report of human biases toward harmonic languages. However, for stronger conclusions, further investigation with more typologically diverse languages and NP perturbations is needed.

\paragraph{Summary} Experiment 3 shows that while LMs do not reflect a gradient of difficulty measured by perplexity in learning different NP orders based on typological prevalence, they may exhibit human-aligned generalization patterns for typologically unattested languages in the generalization test. The differences between perplexity and targeted evaluation results are consistent with \citet{xu2025can}'s findings, which show similar discrepancies.

% Importantly, all five languages perturbed only the four POS tags within the NP structure, which altered adjacent constructions like relative clauses and PPs, potentially rendering the NP structure unanalysable (e.g., \textit{the cute dog that is running}$\rightarrow$\textit{dog the cute that is running} in \textcolor{nnda}{\textsc{\textbf{N}nda}} perturbation). Despite these unnatural NP constructions, language models surprisingly still assigned lower perplexity.

% Languages are never ever random \citep{kilgarriff2005language}.
% \citet{saffran2008grammatical} children learn reverse-word order.

\subsection{Discussion of Perplexity Results}

Why doesn't LM perplexity distinguish between attested and unattested languages? We propose two key factors that influence LM learning outcomes: \textit{randomness} and \textit{constituency structure}. By \textit{randomness}, we refer to whether the perturbation function produces a perturbed text that can be deterministically recovered to its original form. By \textit{constituency structure}, we mean whether the phrase structures of the original language are preserved in the perturbed version.

Regarding randomness, as LMs are simply \textit{distributions over strings} \citep{borenstein-etal-2024-languages}, introducing randomness increases unpredictability of the text, thus increasing the entropy of the sequence. This explains why NP-perturbed unattested languages show lower perplexity than attested languages and \textcolor{random}{\textsc{np\_random}} variants. The reasoning is that our perturbation procedure enforces a strict ordering procedure, which may be (sometimes) violated in the original attested language. For example, although English is a \textcolor{dnan}{\textsc{dnaN}} language, certain constructions such as the \textsc{dann} (\textsc{det-adj-num-noun}; e.g., \textit{a beautiful five days}) does not follow the dominant pattern. Once POS tag orders are normalized within NPs, the resulting constructions become more predictable. Therefore, all normalized NPs, including our unattested NPs, may have lower overall entropy, which could explain why they are easier to learn. In fact, the normalized \textcolor{dnan}{\textsc{dnaN}}, which has the same typical word order as English, shows lower perplexity than the original, unnormalized English; and the same applies to our other languages in this experiment.

Regarding constituency structure, we hypothesize that disrupting constituency weakens local dependency relations within phrase structures. This explains why in experiments 1 and 2, all LMs' perplexities for impossible languages are higher than for NP-perturbed languages, despite maintaining a deterministic order (Figure~\ref{fig:cc}).
Similarly, this may also explain the higher perplexity of count-based grammars in \citet{kallini-etal-2024-mission}, where the authors insert a morphological marker a certain number of words or tokens after a host word. The count-based insertion may disrupt phrase structure integrity.\footnote{We do not replicate these count-based experiments.} 

In sum, this discussion points to a potential confound in our experiments: although the texts are parallel in content, languages with normalized NP structures may have lower entropy. In this case, even if LMs learn all languages equally well, lower entropy would naturally lead to lower perplexity. Future work could control for entropy across NP-perturbed languages to test whether perplexity differences persist.

\section{Discussion \& Conclusion}

In this paper, we extend \citet{kallini-etal-2024-mission} to a broader multilingual context using two new parallel corpora. Our findings complement their work, suggesting that models have a preference for human-like languages, although the preference is somewhat gradient and depends on the testing setup.
First, while GPT-2 small assigns lower perplexity to attested languages compared to their impossible variants, the difference is sometimes not significant, especially later in training.
Second, the model does not fully separate all attested from all unattested or impossible languages, but it \emph{does} generally learn attested languages better, achieving a separability of $0.75$ between the two classes based on perplexity. In the NP word order experiments, some unattested languages exhibit lower perplexity than their attested counterparts, despite having orderings that violate Greenberg's Universal 20. 
However, when assessed using targeted evaluation methods, a more promising pattern emerges:
GPT-2 seems to favor typologically attested, as opposed to unattested NP variants, and shows some preference for harmonic word orderings. \looseness=-1
%However, lower perplexity does not necessarily imply better generalization.

What to make of these results in the context of our original question--whether LMs can serve as cognitive models? While our results show that GPT-2 does not behave as we might expect from a fully human-like learner, they also demonstrate that it has a soft preference for attested over impossible languages. Skeptics have previously linked LMs to a bad theory of physics in which ``anything goes.''\footnote{Chomsky, quoted from an email to Gary Marcus: \textit{``You can’t go to a physics conference and say: I’ve got a great theory.  It accounts for everything and is so simple it can be captured in two words: `Anything goes.' All known and unknown laws of nature are accommodated, no failures.  Of course, everything impossible is accommodated also.''}} In line with \citet{kallini-etal-2024-mission}, our results demonstrate that these models do not instantiate an ``anything goes'' hypothesis. Rather, their incremental data-processing architectures represent a useful starting point for studying human language processing and learning. Refining models to better align with humans is possible and will likely lead to lasting insights about human cognitive architecture.

\section{Limitations}
\label{sec:limitations}
We acknowledge that our experiments rely on GPT-2 Small, which may not generalize to larger models. This choice was made for two reasons: (1) running experiments across multiple languages is computationally expensive; (2) we aimed for comparability with \citet{kallini-etal-2024-mission}. Future work could explore whether our findings hold for larger models or similarly sized models with different architectures. Additionally, the dataset used for training the language model is relatively small. This is a deliberate trade-off between data size and linguistic diversity. While a larger dataset might yield more robust results, our approach ensures broader typological coverage. In our experiments on unattested languages, we generated synthetic data by perturbing languages based on Universal 20. However, linguistic correlations extend beyond word order universals. For instance, Greenbergian correlations \citep{dryer1992greenbergian} suggest that verb-object order often correlates with other features such as adposition-noun phrase order and determiner-noun phrase order. Future work will explore more nuanced perturbations to better capture such cross-linguistic dependencies. Lastly, the data we use has not been manually checked yet. It is possible that our parallel corpora include noise that might influence the learning results. 

\section{Ethics Statement}
%Our research adheres to ethical guidelines in data collection, model development, and evaluation. 
We use publicly available datasets, ensuring that no private or personally identifiable information is included. Our dataset selection prioritizes linguistic diversity while maintaining data transparency. Regarding computational resources, we use GPT-2 small trained on A-100 and V-100 GPUs. Each experiment on each language took around 10-12 hours. 
%Finally, our research is intended for advancing linguistic understanding in computational models and does not facilitate any malicious applications. We encourage responsible usage and open discussions on the ethical implications of NLP research.

\section{Acknowledgments}
We thank Amir Zeldes, Nathan Schneider, Yilun Zhu, Dan DeGenaro, Wesley Scivetti, and all members of PICoL and NERT for their helpful feedback and support.
\bibliography{custom}
\appendix
\newpage
\section{Experiment Results of Replicating \citet{kallini-etal-2024-mission}}
\label{replication}
\begin{figure*}[!th]
    \centering
    \includegraphics[width=0.99\linewidth]{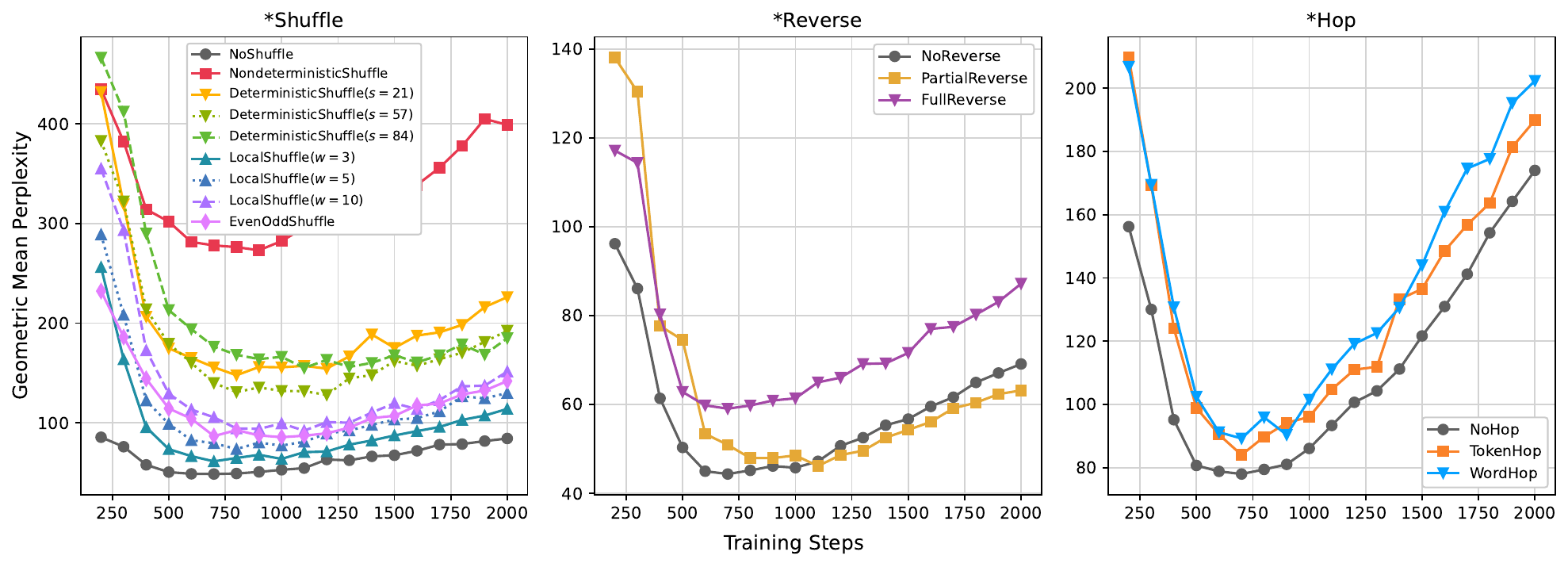}
    \caption{Replication of \citep{kallini-etal-2024-mission} with 10M words from BabyLM Challenge dataset (strict-small track)}
    \label{fig:replication}
\end{figure*}

We implement the training and evaluation following the same experiment setting from \citet{kallini-etal-2024-mission}, only on a 10M word subset of their original data. The result is shown in Figure~\ref{fig:replication}. Unlike in \citet{kallini-etal-2024-mission}, however, we do observe that test-set perplexity does increase towards the end of training, indicating that models are overfitting on our smaller datasets. We note that we do not observe this overfitting behavior in the experiments presented in the main text, where the heldout perplexity continues to decrease (or plateau) throughout training.

We calculate Spearman’s rank correlation between our results for the \textit{*shuffled} languages and those of \citet{kallini-etal-2024-mission} at every 200-step interval from 400 to 1,200. The Spearman’s $\rho$ is consistently \textbf{1} (\textit{p} < 0.001), indicating perfect agreement between the rankings, showing that 10M words are sufficient enough to replicate the language modeling experiments for which \citet{kallini-etal-2024-mission} originally used 100M words.

In experiment 1, we also conducted a Spearman's Ranking Correlation test between the results on OPUS30 English and those from \citet{kallini-etal-2024-mission}'s experiments. We grouped the \textsc{shuffle\_deterministic} languages together and observed that the ranking of our English impossible variants aligns perfectly with that reported by \citet{kallini-etal-2024-mission} ($\rho=1$, $p=0.0027$). 
\begin{figure*}[t]
    \centering
    \includegraphics[width=\linewidth]{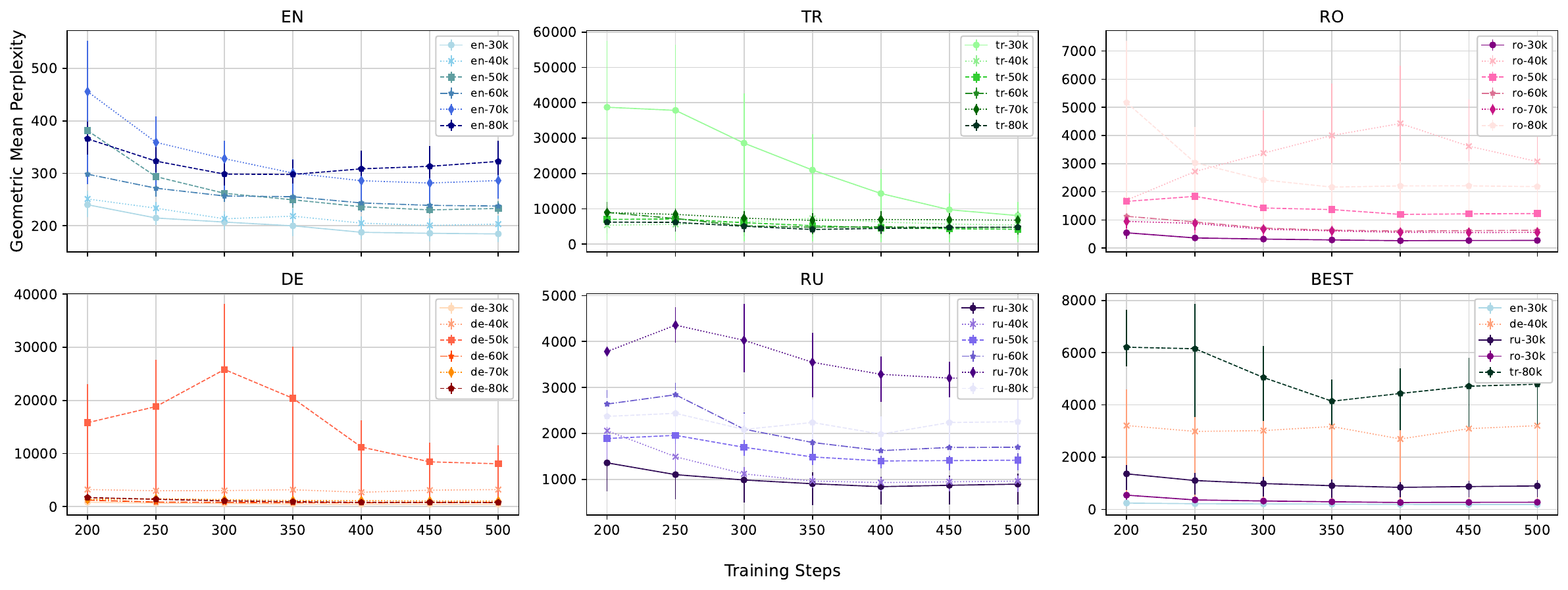}
    \caption{Perplexity results on the development set (10K sentences) for five languages (EN, TR, RO, DE, RU), trained on a 10M-sentence training set across different vocabulary sizes. Error bars represent the first and last quartiles (25\% and 75\%) of the results. A plot for the optimized vocabulary size (labeled ‘BEST’) is also included, showing high variance for TR and RU even with optimized vocabulary size.}
    \label{vairance}
\end{figure*}
\begin{table*}[!th]
\small
    \centering
    \begin{tabular}{p{1.3cm}|p{2.5cm}|p{2.5cm} p{2.5cm} p{2.5cm} p{2.5cm}}
    \toprule
      Language & Treebank  & \multicolumn{4}{c}{POS-tags} \\ 
       \cmidrule(lr){3-6}
      & & \textsc{det} & \textsc{num} & \textsc{adj} & \textsc{noun} \\
    \midrule
       English & Penn Treebank \citep{marcus-etal-1993-building}  & \textsc{dt}, \textsc{prp\$}, \textsc{pdt}, \textsc{pos} 
               & \textsc{qp}, \textsc{\$}, \textsc{cd} 
               & \textsc{rb}, \textsc{adjp}, \textsc{jjr}, \textsc{jjs}, \textsc{jj} 
               & \textsc{nn}, \textsc{nns}, \textsc{nnp}, \textsc{nnps} \\
       Italian & VIT\citep{delmonte2007vit}  & \textsc{det} 
               & \textsc{num}, \textsc{sq} 
               & \textsc{adj}, \textsc{sa} 
               & \textsc{noun}, \textsc{pron}, \textsc{propn}, \textsc{sym}, \textsc{x} \\
       Chinese & CTB 3.0\citep{xue2005penn}  & \textsc{dt}, \textsc{m}, \textsc{clp}, \textsc{dp} 
               & \textsc{cd}, \textsc{od}, \textsc{qp} 
               & \textsc{jj}, \textsc{adjp}, \textsc{dnp}, \textsc{dec}, \textsc{deg} 
               & \textsc{nn}, \textsc{np}, \textsc{nr}, \textsc{nt}, \textsc{prp}, \textsc{pn}, \textsc{fw} \\
       Portuguese & Cintil \citep{barreto-etal-2006-open} & \textsc{det}, \textsc{d}, \textsc{dem}, \textsc{poss}, \textsc{poss'} 
                  & \textsc{qnt}, \textsc{qnt'}, \textsc{num}, \textsc{percentp}, \textsc{percentp'}, \textsc{card}, \textsc{card'} 
                  & \textsc{adj}, \textsc{ap} 
                  & \textsc{n'}, \textsc{noun}, \textsc{pron} \\
    \bottomrule
    \end{tabular}
    \caption{POS-tag categories across languages}
    \label{tab:pos_tags}
\end{table*}

\begin{table}[!h]
\small
    \centering
    \begin{tabular}{c|cccccccccccccccc}
    \toprule
       \textsc{langs}  & \textsc{ar} &\textsc{tr}& \textsc{ru}& \textsc{pl}& \textsc{de}& \textsc{it}  \\
       \textsc{tcw}  & 2.19& 2.05 & 2.05 & 1.98& 1.65 & 1.40\\
       \midrule
      \textsc{langs}   & \textsc{pt} & \textsc{nl} & \textsc{ro} & \textsc{en} & \textsc{fr}\\
     \textsc{tcw}  & 1.68&  1.51 & 1.81 & 1.45& 1.67\\
     \bottomrule
    \end{tabular}
    \caption{TCW per language by each of their pretrained tokenizer}
    \label{tab:ctc}
\end{table}

% \begin{table}[!h]
% \small
%     \centering
%     \begin{tabular}{c|cccccccccccccccc}
%     \toprule
%        \textsc{langs}  & \textsc{ar} &\textsc{tr}& \textsc{ru}& \textsc{pl}& \textsc{de}& \textsc{it}  \\
%        \textsc{tcw}  & 13.25& 11.52 & 13.02 & 12.10& 13.57 & 13.30\\
%        \midrule
%       \textsc{langs}   & \textsc{pt} & \textsc{nl} & \textsc{ro} & \textsc{en} & \textsc{fr}\\
%      \textsc{tcw}  & 13.25&  12.85 &14.79 & 13.32& 14.71\\
%      \bottomrule
%     \end{tabular}
%     \caption{CTC per language per sentence by each of their pretrained tokenizer}
%     \label{tab:realctc}
% \end{table}

% \begin{table}[!h]
% \small
%     \centering
%     \begin{tabular}{c|cccccccccccccccc}
%     \toprule
%        \textsc{langs}  & \textsc{ar} &\textsc{tr}& \textsc{ru}& \textsc{pl}& \textsc{de}& \textsc{it}  \\
%        \textsc{ctc}  & 16.41& 16.27 & 16.39 & 16.31& 16.43 & 16.41\\
%        \midrule
%       \textsc{langs}   & \textsc{pt} & \textsc{nl} & \textsc{ro} & \textsc{en} & \textsc{fr}\\
%      \textsc{ctc}  & 16.41& 16.37 &16.52 & 16.41& 16.51\\
%      \bottomrule
%     \end{tabular}
%     \caption{$log(CTC)$ per language by each of their pretrained tokenizer}
%     \label{tab:logctc}
% \end{table}

\section{Tokenization Pilot Experiments and Results}
\label{pilot_study}

In our experiments, where we trained tokenizers for each language using 10M words (around 60MB data), testing vocabulary sizes ranging from 30K to 80K in increments of 10K, we observed two key findings: (1) Tokenizers trained with around 60MB data resulted in unstable language modeling outcomes, and (2) different languages require distinct optimal vocabulary sizes: morphologically richer languages tend to have a larger vocabulary size. We also observed that even when trained on the corpus with matching content, not all languages are equally learnable in terms of their perplexity. These results are shown in Figure~\ref{vairance}. Additionally, agglutinative languages like Turkish, with their large number of unique tokens, made large vocabulary sizes impractical. For instance, Turkish has three times the number of unique words as English (467K vs. 140K), and applying $0.4 \times |V|$ would result in a vocabulary size of 186K, which is too large for efficient language model training with the limited data available and a small model.

\begin{table*}[th]
\small
    \centering
    \begin{tabular}{lllll}
    \toprule
       Language   & Family                    & Word Order           & Morphology                          \\
    \midrule
    \textbf{\textit{OPUS12}}\\
    English    & Indo-European (Germanic)    & SVO                  & Analytic                            \\
    German     & Indo-European (Germanic)    & No dominant          & Fusional                            \\
    Russian    & Indo-European (Slavonic)    & SVO                  & Fusional                            \\
    Romanian   & Indo-European (Romance)     & SVO                  & Fusional                            \\
     Turkish    & Turkic (Altaic)             & SOV                  & Agglutinative                       \\
    Dutch      & Indo-European (Germanic)    & No dominant          & Fusional                            \\
    
    Polish     & Indo-European (Slavonic)    & SVO                  & Fusional                            \\    
    Portuguese & Indo-European (Romance)     & SVO                  & Fusional                            \\
    Italian    & Indo-European (Romance)     & SVO                  & Fusional                            \\
    French     & Indo-European (Romance)     & SVO                  & Fusional                            \\
    
    Chinese    & Sino-Tibetan                & SVO                  & Analytic                            \\
    Arabic     & Afro-Asiatic (Semitic)      & VSO                  & Root-based (nonconcatenative)       \\
   
    \midrule
    \textbf{\textit{OPUS30}}\\
    Spanish & Indo-European (Romance) & SVO & Fusional\\
    Czech      & Indo-European (Slavonic)    & SVO                  & Fusional  \\
    Bulgarian  & Indo-European (Slavonic)    & SVO                  & Fusional                            \\
     Slovak     & Indo-European (Slavonic)    & SVO                  & Fusional                            \\
    Serbian    & Indo-European (Slavonic)    & SVO                  & Fusional                            \\
    Croatian   & Indo-European (Slavonic)    & SVO                  & Fusional                            \\
    Ukrainian  & Indo-European (Slavonic)    & SVO                  & Fusional                            \\
    Danish     & Indo-European (Germanic)    & SVO                  & Fusional                            \\
    Swedish    & Indo-European (Germanic)    & SVO                  & Fusional                            \\
    Greek      & Indo-European (Hellenic)    & No dominant                 & Fusional                            \\
    Persian    & Indo-European (Indo-Iranian) & SVO                 & Fusional                            \\
       Lithuanian & Indo-European (Baltic)      & SVO                  & Fusional                            \\
    Vietnamese & Austroasiatic               & SVO                  & Analytic                            \\
    Hebrew     & Afro-Asiatic (Semitic)      & VSO                  & Root-based (nonconcatenative)       \\
    
    Hungarian  & Uralic                      & SVO                  & Agglutinative                       \\
    Indonesian & Austronesian                & SVO                  & Analytic                            \\
    Japanese   & Japonic                     & SOV                  & Agglutinative                       \\
    Korean     & Koreanic                    & SOV                  & Agglutinative                       \\
 
    % \midrule
    \bottomrule
    \end{tabular}
    \caption{Typological features of the OPUS12 and OPUS30 corpora, with OPUS30 including 18 additional languages beyond those in OPUS12.}
    \label{tab:language}
\end{table*}

\section{Details of OPUS12 and OPUS30}\label{appendix:data-details}
The typological features of languages used in the two corpora are reported in Table~\ref{tab:language}. The licensing terms vary depending on their original sources, listed below.
\begin{itemize}
    \item NLLB: \href{https://opendatacommons.org/licenses/by/1-0/}{ODC-By}
    \item TED2020: \href{https://creativecommons.org/licenses/by-nc-nd/4.0/deed.en}{CC BY–NC–ND 4.0 International}; for details, see \href{https://www.ted.com/about/our-organization/our-policies-terms/ted-talks-usage-policy}{the official website}.
    \item Bible: \href{https://creativecommons.org/publicdomain/zero/1.0/deed.en}{CC0 1.0}
    \item OpenSubtitles: \href{https://www.gnu.org/licenses/gpl-3.0.en.html}{GNU General Public License v3.0}
    \item MultiCCAligned: unknown; see \href{https://paperswithcode.com/dataset/ccaligned}{the official website}.
\end{itemize}

\section{Tokenizers}
\label{tokenizers}
Table~\ref{tab:tokenizers} shows the details of the tokenizers we use in the experiments. When the training data for a tokenizer is unspecified, we assume it matches the training data used for the corresponding pretrained model. 

\section{TCW \& CTC}
\label{tcw}
The TCW is reported in Table~\ref{tab:ctc}. We use it to measure the morphological richness of a language.
\section{POS tags of each treebank}
\label{treebank}
Different constituency parsers are trained with different treebanks. We select POS-tags that are relevant to the four word classes. The detailed POS-tags for each language can be found in Table~\ref{tab:pos_tags}. 

\begin{table*}[t]
\small 
    \centering
    \begin{tabular}{lrrrr}
    \toprule
    Language & |Vocab| & |Training| & Reference  & Domain \\
    \midrule
    Arabic\footnotemark[1] &64,000 &77GB  & {\tiny\citet{antoun-etal-2021-aragpt2}} &  Web Crawl, Wikipedia, News\\
    Turkish\footnotemark[2] & 50,257 & 100GB & {\tiny \citet{Kesgin_2024}} & Web Crawl, books, news, others \\
    Russian\footnotemark[3] & 50,257 & 450GB & {\tiny \citet{zmitrovich-etal-2024-family}} &  Wikipedia, books, news, books, Web Crawl, Subtitles
    \\        
    Polish\footnotemark[4] & 50,257 & 47GB &\tiny{\citet{papuGaPT2}} & Web Crawl\\
    German\footnotemark[5]  & 50,304 & 156GB & {\tiny\citet{ostendorff2023gpt2wechsel}} & Web Crawl \\
    Italian\footnotemark[6] & 50,176 & Trillions toks & {\tiny\citet{igeniusai2024italia}} & public sources, synthetic data, and domain-specific content\\
    Portugese\footnotemark[7] & 50,258 & 35B tokens & \tiny\citet{lopes-etal-2024-gloria} & Web Crawl, News, Subtitles, EuroParl\\
    Dutch\footnotemark[8]  & 50,257 & 151GB & {\tiny \citet{havinga2023gptneo}} & Web Crawl \\
    Romanian\footnotemark[9] & 64,000 & 40GB & {\tiny\citet{dumitrescu2024gptneo}}  & Web Crawl, Opus, Wikipedia \\
    English\footnotemark[10] & 50,257 & 40GB & {\tiny \citet{radford2019language}} & Web Crawl\\
    French\footnotemark[11]  & 50,262 & 130GB & {\tiny \citet{launay-etal-2022-pagnol}} & Web Crawl \\
    Chinese\footnotemark[12] & 21,128 & 300GB & {\tiny\citet{devlin-etal-2019-bert}} & Wikipedia \\
    \bottomrule
    \end{tabular}
    \caption{Tokenizers, vocabulary sizes, training data sizes, references, pretrained model name, and training data domains for each language tested in our experiments.}
    \label{tab:tokenizers}
\end{table*}
\footnotetext[1]{\footnotesize{\url{https://huggingface.co/aubmindlab/aragpt2-base}}}
\footnotetext[2]{\footnotesize{\url{https://huggingface.co/ytu-ce-cosmos/turkish-gpt2}}}
\footnotetext[3]{\footnotesize{\url{https://huggingface.co/ai-forever/rugpt3large_based_on_gpt2}}}
\footnotetext[4]{\footnotesize{\url{https://huggingface.co/flax-community/papuGaPT2}}}
\footnotetext[5]{\footnotesize{\url{https://huggingface.co/malteos/gpt2-xl-wechsel-german}}}
\footnotetext[6]{\footnotesize{\url{https://huggingface.co/iGeniusAI/Italia-9B-Instruct-v0.1}}}
\footnotetext[7]{\footnotesize{\url{https://huggingface.co/NOVA-vision-language/GlorIA-1.3B}}}
\footnotetext[8]{\footnotesize{\url{https://huggingface.co/yhavinga/gpt-neo-125M-dutch}}}
\footnotetext[9]{\footnotesize{\url{https://huggingface.co/dumitrescustefan/gpt-neo-romanian-780m}}}
\footnotetext[10]{\footnotesize{\url{https://huggingface.co/openai-community/gpt2}}}
\footnotetext[11]{\footnotesize{\url{https://huggingface.co/lightonai/pagnol-xl}}}
\footnotetext[12]{\footnotesize{\url{https://huggingface.co/google-bert/bert-base-chinese}}}

\section{Statistical test between impossible languages}
\label{stats_test}
We conducted Welch's paired t-test comparing different perturbations with shuffle\_control across 12 checkpoints. The results are ordered alphabetically.

We find that for Dutch, Russian, and Turkish, the difference between \textsc{shuffle\_control} and other perturbations is always significant; by contrast, for languages including Arabic, Chinese, English, German, and Romanian, the difference becomes less significant or insignificant in the locally shuffled variants.
\begin{table*}[!h]
\centering
\tiny
\begin{tabular}{lllllllllllll}
\toprule
Perturbation | Step & 100 & 200 & 300 & 400 & 500 & 600 & 700 & 800 & 900 & 1000 & 1100 & 1200\\
\midrule
perturb\_reverse\_full\_word & <0.001 & <0.001 & 0.0036 & 0.0484 & <0.001 & 0.0283 & 0.0781 & 0.5293 & 1 & 0.8811 & 1 & 1\\
shuffle\_deterministic21 & 1 & <0.001 & <0.001 & <0.001 & <0.001 & <0.001 & <0.001 & <0.001 & <0.001 & <0.001 & <0.001 & <0.001\\
shuffle\_deterministic57 & <0.001 & <0.001 & <0.001 & <0.001 & <0.001 & <0.001 & <0.001 & <0.001 & <0.001 & <0.001 & <0.001 & <0.001\\
shuffle\_deterministic84 & 0.7871 & <0.001 & <0.001 & <0.001 & <0.001 & <0.001 & <0.001 & <0.001 & <0.001 & <0.001 & <0.001 & <0.001\\
shuffle\_even\_odd & <0.001 & <0.001 & <0.001 & <0.001 & <0.001 & <0.001 & <0.001 & <0.001 & <0.001 & <0.001 & <0.001 & <0.001\\
\addlinespace
shuffle\_local10 & 0.4242 & <0.001 & <0.001 & <0.001 & <0.001 & <0.001 & <0.001 & <0.001 & <0.001 & <0.001 & <0.001 & <0.001\\
shuffle\_local2 & 1 & <0.001 & <0.001 & <0.001 & <0.001 & 0.7469 & 1 & 1 & 1 & 1 & 1 & 1\\
shuffle\_local3 & 1 & <0.001 & <0.001 & <0.001 & <0.001 & 0.0075 & 0.0062 & 0.1182 & 1 & 1 & 1 & 1\\
shuffle\_local5 & 0.4024 & <0.001 & <0.001 & <0.001 & <0.001 & <0.001 & <0.001 & <0.001 & 0.3782 & 0.3208 & 1 & 1\\
\bottomrule
\end{tabular}
\caption{Welch’s t-test comparing each perturbation with shuffle\_control across 12 checkpoints for \textbf{Arabic}, with Bonferroni adjustment.}
\end{table*}

\begin{table*}[!h]
\centering
\tiny
\begin{tabular}{lllllllllllll}
\toprule
Perturbation | Step & 100 & 200 & 300 & 400 & 500 & 600 & 700 & 800 & 900 & 1000 & 1100 & 1200\\
\midrule
perturb\_reverse\_full\_word & <0.001 & <0.001 & <0.001 & 1 & <0.001 & <0.001 & <0.001 & <0.001 & <0.001 & <0.001 & <0.001 & <0.001\\
shuffle\_deterministic21 & <0.001 & <0.001 & <0.001 & <0.001 & <0.001 & <0.001 & <0.001 & <0.001 & <0.001 & <0.001 & <0.001 & <0.001\\
shuffle\_deterministic57 & <0.001 & <0.001 & <0.001 & <0.001 & <0.001 & <0.001 & <0.001 & <0.001 & <0.001 & <0.001 & <0.001 & <0.001\\
shuffle\_deterministic84 & <0.001 & <0.001 & <0.001 & <0.001 & <0.001 & <0.001 & <0.001 & <0.001 & <0.001 & <0.001 & <0.001 & <0.001\\
shuffle\_even\_odd & <0.001 & <0.001 & <0.001 & <0.001 & <0.001 & <0.001 & <0.001 & <0.001 & <0.001 & <0.001 & <0.001 & <0.001\\
\addlinespace
shuffle\_local10 & <0.001 & <0.001 & <0.001 & <0.001 & <0.001 & <0.001 & <0.001 & <0.001 & <0.001 & <0.001 & <0.001 & <0.001\\
shuffle\_local2 & <0.001 & <0.001 & <0.001 & <0.001 & <0.001 & 0.0142 & 1 & 1 & 1 & 1 & 1 & 1\\
shuffle\_local3 & <0.001 & <0.001 & <0.001 & <0.001 & <0.001 & <0.001 & <0.001 & <0.001 & 0.0012 & 0.0014 & 0.0063 & 0.135\\
shuffle\_local5 & <0.001 & <0.001 & <0.001 & <0.001 & <0.001 & <0.001 & <0.001 & 1 & 1 & 0.6033 & 0.0391 & 0.0016\\
\bottomrule
\end{tabular}
\caption{Welch’s t-test comparing each perturbation with shuffle\_control across 12 checkpoints for \textbf{Chinese}, with Bonferroni adjustment.}
\end{table*}

\begin{table*}[!h]
\centering
\tiny
\begin{tabular}{lllllllllllll}
\toprule
Perturbation | Step & 100 & 200 & 300 & 400 & 500 & 600 & 700 & 800 & 900 & 1000 & 1100 & 1200\\
\midrule
perturb\_reverse\_full\_word & <0.001 & <0.001 & 0.0012 & 0.1697 & 0.0396 & 0.0137 & 0.003 & <0.001 & <0.001 & <0.001 & <0.001 & <0.001\\
shuffle\_deterministic21 & <0.001 & <0.001 & <0.001 & <0.001 & <0.001 & <0.001 & <0.001 & <0.001 & <0.001 & <0.001 & <0.001 & <0.001\\
shuffle\_deterministic57 & 1 & <0.001 & <0.001 & <0.001 & <0.001 & <0.001 & <0.001 & <0.001 & <0.001 & <0.001 & <0.001 & <0.001\\
shuffle\_deterministic84 & <0.001 & <0.001 & <0.001 & <0.001 & <0.001 & <0.001 & <0.001 & <0.001 & <0.001 & <0.001 & <0.001 & <0.001\\
shuffle\_even\_odd & <0.001 & <0.001 & <0.001 & <0.001 & <0.001 & <0.001 & <0.001 & <0.001 & <0.001 & <0.001 & <0.001 & <0.001\\
\addlinespace
shuffle\_local10 & <0.001 & <0.001 & <0.001 & <0.001 & <0.001 & <0.001 & <0.001 & <0.001 & <0.001 & <0.001 & <0.001 & <0.001\\
shuffle\_local2 & <0.001 & <0.001 & 1 & <0.001 & 0.0354 & 0.0277 & 0.0051 & 0.0059 & 0.0158 & <0.001 & 0.0043 & <0.001\\
shuffle\_local3 & <0.001 & <0.001 & <0.001 & <0.001 & <0.001 & <0.001 & <0.001 & <0.001 & <0.001 & 0.002 & 0.0296 & 0.0111\\
shuffle\_local5 & <0.001 & <0.001 & <0.001 & <0.001 & <0.001 & <0.001 & <0.001 & <0.001 & 0.0018 & <0.001 & 0.0049 & 0.0026\\
\bottomrule
\end{tabular}
\caption{Welch’s t-test comparing each perturbation with shuffle\_control across 12 checkpoints for \textbf{Dutch}, with Bonferroni adjustment.}
\end{table*}

\begin{table*}[!h]
\centering
\tiny
\begin{tabular}{lllllllllllll}
\toprule
Perturbation | Step & 100 & 200 & 300 & 400 & 500 & 600 & 700 & 800 & 900 & 1000 & 1100 & 1200\\
\midrule
perturb\_reverse\_full\_word & <0.001 & <0.001 & 0.0188 & 0.0016 & 0.0025 & 0.0751 & 0.0028 & 0.684 & 0.938 & 1 & 0.8737 & 1\\
shuffle\_deterministic21 & <0.001 & <0.001 & <0.001 & <0.001 & <0.001 & <0.001 & <0.001 & <0.001 & <0.001 & <0.001 & <0.001 & <0.001\\
shuffle\_deterministic57 & <0.001 & <0.001 & <0.001 & <0.001 & <0.001 & <0.001 & <0.001 & <0.001 & <0.001 & <0.001 & <0.001 & <0.001\\
shuffle\_deterministic84 & <0.001 & <0.001 & <0.001 & <0.001 & <0.001 & <0.001 & <0.001 & <0.001 & <0.001 & <0.001 & <0.001 & <0.001\\
shuffle\_even\_odd & <0.001 & <0.001 & <0.001 & 0.0078 & <0.001 & <0.001 & 0.0022 & 0.4015 & 0.0068 & 0.0013 & 0.0235 & 0.0812\\
shuffle\_local10 & <0.001 & <0.001 & <0.001 & <0.001 & <0.001 & <0.001 & <0.001 & <0.001 & <0.001 & <0.001 & <0.001 & <0.001\\
shuffle\_local2 & 0.0458 & <0.001 & <0.001 & 0.0055 & 0.0251 & 1 & 1 & 1 & 1 & 1 & 1 & 1\\
shuffle\_local3 & <0.001 & <0.001 & <0.001 & <0.001 & <0.001 & <0.001 & <0.001 & <0.001 & <0.001 & <0.001 & 0.0269 & 0.0878\\
shuffle\_local5 & <0.001 & <0.001 & <0.001 & <0.001 & <0.001 & <0.001 & <0.001 & <0.001 & <0.001 & <0.001 & 0.0167 & 0.1072\\
\bottomrule
\end{tabular}
\caption{Welch’s t-test comparing each perturbation with shuffle\_control across 12 checkpoints for \textbf{English}, with Bonferroni adjustment.}
\end{table*}

\begin{table*}[!h]
\centering
\tiny
\begin{tabular}{lllllllllllll}
\toprule
Perturbation | Step & 100 & 200 & 300 & 400 & 500 & 600 & 700 & 800 & 900 & 1000 & 1100 & 1200\\
\midrule
perturb\_reverse\_full\_word & <0.001 & 0.0284 & 1 & 0.008 & 1 & 1 & 1 & 1 & 1 & 1 & 1 & 1\\
shuffle\_deterministic21 & <0.001 & <0.001 & <0.001 & <0.001 & <0.001 & 0.0046 & 0.6694 & 1 & 1 & 1 & 1 & 1\\
shuffle\_deterministic57 & 1 & <0.001 & <0.001 & <0.001 & <0.001 & 0.0066 & 0.6324 & 1 & 1 & 1 & 1 & 1\\
shuffle\_deterministic84 & <0.001 & 0.0203 & <0.001 & <0.001 & <0.001 & <0.001 & 0.0402 & 1 & 1 & 1 & 1 & 1\\
shuffle\_even\_odd & <0.001 & <0.001 & 1 & <0.001 & 1 & 1 & 1 & 1 & 1 & 1 & 1 & 1\\
\addlinespace
shuffle\_local10 & <0.001 & <0.001 & 0.3162 & <0.001 & 1 & 1 & 1 & 1 & 1 & 1 & 1 & 1\\
shuffle\_local2 & <0.001 & <0.001 & 1 & 1 & 1 & 1 & 1 & 1 & 1 & 1 & 1 & 1\\
shuffle\_local3 & <0.001 & <0.001 & 1 & 0.5046 & 1 & 1 & 1 & 1 & 1 & 1 & 1 & 1\\
shuffle\_local5 & <0.001 & <0.001 & 1 & 0.0257 & 1 & 1 & 1 & 1 & 1 & 1 & 1 & 1\\
\bottomrule
\end{tabular}
\caption{Welch’s t-test comparing each perturbation with shuffle\_control across 12 checkpoints for \textbf{French}, with Bonferroni adjustment.}
\end{table*}

\begin{table*}[!h]
\centering
\tiny
\begin{tabular}{lllllllllllll}
\toprule
Perturbation | Step & 100 & 200 & 300 & 400 & 500 & 600 & 700 & 800 & 900 & 1000 & 1100 & 1200\\
\midrule
perturb\_reverse\_full\_word & <0.001 & <0.001 & <0.001 & <0.001 & <0.001 & <0.001 & <0.001 & <0.001 & <0.001 & <0.001 & <0.001 & <0.001\\
shuffle\_deterministic21 & <0.001 & <0.001 & <0.001 & <0.001 & <0.001 & <0.001 & <0.001 & <0.001 & <0.001 & <0.001 & <0.001 & <0.001\\
shuffle\_deterministic57 & <0.001 & <0.001 & <0.001 & <0.001 & <0.001 & <0.001 & <0.001 & <0.001 & <0.001 & <0.001 & <0.001 & <0.001\\
shuffle\_deterministic84 & <0.001 & <0.001 & <0.001 & <0.001 & <0.001 & <0.001 & <0.001 & <0.001 & <0.001 & <0.001 & <0.001 & <0.001\\
shuffle\_even\_odd & <0.001 & <0.001 & <0.001 & <0.001 & <0.001 & <0.001 & <0.001 & <0.001 & <0.001 & <0.001 & <0.001 & <0.001\\
\addlinespace
shuffle\_local10 & <0.001 & <0.001 & <0.001 & <0.001 & <0.001 & <0.001 & <0.001 & <0.001 & <0.001 & <0.001 & <0.001 & <0.001\\
shuffle\_local2 & <0.001 & <0.001 & <0.001 & <0.001 & 0.034 & 0.0974 & 0.4926 & 0.4579 & 1 & 0.2287 & 1 & 0.4976\\
shuffle\_local3 & <0.001 & <0.001 & <0.001 & <0.001 & <0.001 & <0.001 & 0.0035 & 0.0185 & 0.011 & 0.0675 & 0.14 & 0.2177\\
shuffle\_local5 & <0.001 & <0.001 & <0.001 & <0.001 & <0.001 & <0.001 & <0.001 & 0.001 & <0.001 & <0.001 & <0.001 & 0.0017\\
\bottomrule
\end{tabular}
\caption{Welch’s t-test comparing each perturbation with shuffle\_control across 12 checkpoints for \textbf{German}, with Bonferroni adjustment.}
\end{table*}

\begin{table*}[!h]
\centering
\tiny
\begin{tabular}{lllllllllllll}
\toprule
Perturbation | Step & 100 & 200 & 300 & 400 & 500 & 600 & 700 & 800 & 900 & 1000 & 1100 & 1200\\
\midrule
perturb\_reverse\_full\_word & <0.001 & 1 & 1 & 1 & 1 & 1 & 1 & 1 & 1 & 1 & 1 & 1\\
shuffle\_deterministic21 & <0.001 & <0.001 & 1 & 0.5567 & 1 & 1 & 1 & 1 & 1 & 1 & 1 & 1\\
shuffle\_deterministic57 & 0.002 & 0.0107 & 0.8726 & 0.3123 & 1 & 1 & 1 & 1 & 1 & 1 & 1 & 1\\
shuffle\_deterministic84 & <0.001 & <0.001 & 0.0112 & 0.0013 & 0.002 & 1 & 1 & 1 & 0.8841 & 1 & 1 & 1\\
shuffle\_even\_odd & <0.001 & <0.001 & 0.1087 & 0.3001 & 0.1957 & 1 & 1 & 1 & 1 & 1 & 1 & 1\\
\addlinespace
shuffle\_local\_word3 & <0.001 & 0.0846 & 1 & 1 & 1 & 1 & 1 & 1 & 1 & 1 & 1 & 1\\
shuffle\_local10 & 0.0098 & <0.001 & <0.001 & 0.4004 & 1 & 1 & 1 & 1 & 1 & 1 & 1 & 1\\
shuffle\_local2 & <0.001 & 1 & 1 & 0.3393 & 0.0606 & 0.1196 & 0.1088 & 0.1105 & 0.1625 & 0.1634 & 0.2567 & 0.2097\\
shuffle\_local3 & 1 & 1 & 1 & 1 & 1 & 1 & 1 & 1 & 1 & 1 & 1 & 1\\
shuffle\_local5 & 1 & 1 & 1 & 1 & 1 & 1 & 1 & 1 & 1 & 1 & 1 & 1\\
\bottomrule
\end{tabular}
\caption{Welch’s t-test comparing each perturbation with shuffle\_control across 12 checkpoints for \textbf{Italian}, with Bonferroni adjustment.}
\end{table*}

\begin{table*}[!h]
\centering
\tiny
\begin{tabular}{lllllllllllll}
\toprule
Perturbation | Step & 100 & 200 & 300 & 400 & 500 & 600 & 700 & 800 & 900 & 1000 & 1100 & 1200\\
\midrule
perturb\_reverse\_full\_word & <0.001 & <0.001 & <0.001 & <0.001 & <0.001 & <0.001 & <0.001 & <0.001 & <0.001 & <0.001 & <0.001 & <0.001\\
shuffle\_deterministic21 & 0.0018 & <0.001 & <0.001 & <0.001 & <0.001 & <0.001 & <0.001 & <0.001 & <0.001 & <0.001 & <0.001 & <0.001\\
shuffle\_deterministic57 & <0.001 & <0.001 & <0.001 & <0.001 & <0.001 & <0.001 & <0.001 & <0.001 & <0.001 & <0.001 & <0.001 & <0.001\\
shuffle\_deterministic84 & <0.001 & <0.001 & <0.001 & <0.001 & <0.001 & <0.001 & <0.001 & <0.001 & <0.001 & <0.001 & <0.001 & <0.001\\
shuffle\_even\_odd & 0.0049 & <0.001 & <0.001 & <0.001 & <0.001 & <0.001 & <0.001 & <0.001 & <0.001 & <0.001 & <0.001 & <0.001\\
\addlinespace
shuffle\_local10 & <0.001 & <0.001 & <0.001 & <0.001 & <0.001 & <0.001 & <0.001 & <0.001 & <0.001 & <0.001 & <0.001 & <0.001\\
shuffle\_local2 & 0.0288 & <0.001 & <0.001 & <0.001 & <0.001 & 0.0166 & 0.0577 & 0.1104 & 0.0993 & 0.0446 & 0.0521 & 0.0508\\
shuffle\_local3 & <0.001 & <0.001 & <0.001 & <0.001 & <0.001 & <0.001 & <0.001 & <0.001 & <0.001 & <0.001 & <0.001 & <0.001\\
shuffle\_local5 & <0.001 & <0.001 & <0.001 & <0.001 & <0.001 & <0.001 & <0.001 & <0.001 & <0.001 & <0.001 & <0.001 & <0.001\\
\bottomrule
\end{tabular}
\caption{Welch’s t-test comparing each perturbation with shuffle\_control across 12 checkpoints for \textbf{Polish}, with Bonferroni adjustment.}
\end{table*}

\begin{table*}[!h]
\centering
\tiny
\begin{tabular}{lllllllllllll}
\toprule
Perturbation | Step & 100 & 200 & 300 & 400 & 500 & 600 & 700 & 800 & 900 & 1000 & 1100 & 1200\\
\midrule
perturb\_reverse\_full\_word & <0.001 & <0.001 & <0.001 & <0.001 & <0.001 & 0.0056 & 0.0028 & 0.0659 & 0.1986 & 1 & 1 & 1\\
shuffle\_deterministic21 & 1 & <0.001 & <0.001 & <0.001 & <0.001 & <0.001 & <0.001 & <0.001 & <0.001 & 0.0358 & 0.0039 & 0.0778\\
shuffle\_deterministic57 & 1 & <0.001 & <0.001 & <0.001 & <0.001 & <0.001 & <0.001 & <0.001 & <0.001 & 1 & 0.4359 & 1\\
shuffle\_deterministic84 & <0.001 & <0.001 & <0.001 & <0.001 & <0.001 & <0.001 & <0.001 & <0.001 & <0.001 & 0.006 & 0.0012 & 0.0137\\
shuffle\_even\_odd & 1 & <0.001 & <0.001 & 0.0569 & <0.001 & 0.3289 & 1 & 1 & 1 & 1 & 1 & 1\\
\addlinespace
shuffle\_local10 & 1 & <0.001 & <0.001 & <0.001 & <0.001 & <0.001 & <0.001 & 0.1379 & 1 & 1 & 1 & 1\\
shuffle\_local2 & 1 & 1 & <0.001 & 1 & 0.1401 & 1 & 1 & 1 & 1 & 1 & 1 & 1\\
shuffle\_local3 & 0.8382 & <0.001 & <0.001 & 0.0189 & <0.001 & 1 & 1 & 1 & 1 & 1 & 1 & 1\\
shuffle\_local5 & 1 & <0.001 & <0.001 & 0.6868 & <0.001 & 1 & 1 & 1 & 1 & 1 & 1 & 1\\
\bottomrule
\end{tabular}
\caption{Welch’s t-test comparing each perturbation with shuffle\_control across 12 checkpoints for \textbf{Portuguese}, with Bonferroni adjustment.}
\end{table*}

\begin{table*}[!h]
\centering
\tiny
\begin{tabular}{lllllllllllll}
\toprule
Perturbation | Step & 100 & 200 & 300 & 400 & 500 & 600 & 700 & 800 & 900 & 1000 & 1100 & 1200\\
\midrule
perturb\_reverse\_full\_word & <0.001 & <0.001 & <0.001 & <0.001 & <0.001 & <0.001 & 1 & <0.001 & <0.001 & 0.0693 & 0.0291 & 0.0033\\
shuffle\_deterministic21 & <0.001 & <0.001 & <0.001 & <0.001 & <0.001 & <0.001 & 0.6784 & <0.001 & <0.001 & <0.001 & <0.001 & <0.001\\
shuffle\_deterministic57 & <0.001 & <0.001 & <0.001 & <0.001 & <0.001 & <0.001 & 1 & <0.001 & <0.001 & <0.001 & <0.001 & <0.001\\
shuffle\_deterministic84 & <0.001 & <0.001 & <0.001 & <0.001 & <0.001 & <0.001 & 0.1518 & <0.001 & <0.001 & <0.001 & <0.001 & <0.001\\
shuffle\_even\_odd & <0.001 & <0.001 & <0.001 & <0.001 & <0.001 & <0.001 & 1 & <0.001 & <0.001 & 0.0059 & 0.011 & 0.1607\\
\addlinespace
shuffle\_local10 & <0.001 & <0.001 & <0.001 & <0.001 & <0.001 & <0.001 & 1 & <0.001 & <0.001 & <0.001 & <0.001 & <0.001\\
shuffle\_local2 & <0.001 & <0.001 & <0.001 & <0.001 & <0.001 & 0.0311 & 1 & 0.1407 & 0.123 & 0.0092 & 0.0079 & 0.0246\\
shuffle\_local3 & <0.001 & <0.001 & <0.001 & <0.001 & <0.001 & <0.001 & 1 & <0.001 & <0.001 & <0.001 & <0.001 & 0.0292\\
shuffle\_local5 & <0.001 & <0.001 & <0.001 & <0.001 & <0.001 & <0.001 & 1 & <0.001 & <0.001 & 0.0014 & <0.001 & 0.2879\\
\bottomrule
\end{tabular}
\caption{Welch’s t-test comparing each perturbation with shuffle\_control across 12 checkpoints for \textbf{Romanian}, with Bonferroni adjustment.}
\end{table*}

\begin{table*}[!h]
\centering
\tiny
\begin{tabular}{lllllllllllll}
\toprule
Perturbation | Step & 100 & 200 & 300 & 400 & 500 & 600 & 700 & 800 & 900 & 1000 & 1100 & 1200\\
\midrule
perturb\_reverse\_full\_word & <0.001 & <0.001 & <0.001 & <0.001 & <0.001 & <0.001 & <0.001 & <0.001 & <0.001 & <0.001 & <0.001 & <0.001\\
shuffle\_deterministic21 & <0.001 & <0.001 & <0.001 & <0.001 & <0.001 & <0.001 & <0.001 & <0.001 & <0.001 & <0.001 & <0.001 & <0.001\\
shuffle\_deterministic57 & <0.001 & <0.001 & <0.001 & <0.001 & <0.001 & <0.001 & <0.001 & <0.001 & <0.001 & <0.001 & <0.001 & <0.001\\
shuffle\_deterministic84 & <0.001 & <0.001 & <0.001 & <0.001 & <0.001 & <0.001 & <0.001 & <0.001 & <0.001 & <0.001 & <0.001 & <0.001\\
shuffle\_even\_odd & <0.001 & <0.001 & <0.001 & <0.001 & <0.001 & <0.001 & <0.001 & <0.001 & <0.001 & <0.001 & <0.001 & <0.001\\
\addlinespace
shuffle\_local10 & <0.001 & <0.001 & <0.001 & <0.001 & <0.001 & <0.001 & <0.001 & <0.001 & <0.001 & <0.001 & <0.001 & <0.001\\
shuffle\_local2 & <0.001 & <0.001 & <0.001 & 1 & <0.001 & <0.001 & <0.001 & <0.001 & <0.001 & <0.001 & <0.001 & <0.001\\
shuffle\_local3 & <0.001 & <0.001 & <0.001 & <0.001 & <0.001 & <0.001 & <0.001 & <0.001 & <0.001 & <0.001 & <0.001 & <0.001\\
shuffle\_local5 & <0.001 & <0.001 & <0.001 & <0.001 & <0.001 & <0.001 & <0.001 & <0.001 & <0.001 & <0.001 & <0.001 & <0.001\\
\bottomrule
\end{tabular}
\caption{Welch’s t-test comparing each perturbation with shuffle\_control across 12 checkpoints for \textbf{Russian}, with Bonferroni adjustment.}
\end{table*}

\begin{table*}[!h]
\centering
\tiny
\begin{tabular}{lllllllllllll}
\toprule
Perturbation | Step & 100 & 200 & 300 & 400 & 500 & 600 & 700 & 800 & 900 & 1000 & 1100 & 1200\\
\midrule
perturb\_reverse\_full\_word & <0.001 & <0.001 & <0.001 & <0.001 & <0.001 & <0.001 & <0.001 & <0.001 & <0.001 & <0.001 & <0.001 & <0.001\\
shuffle\_deterministic21 & 0.0551 & <0.001 & <0.001 & <0.001 & <0.001 & <0.001 & <0.001 & <0.001 & <0.001 & <0.001 & <0.001 & <0.001\\
shuffle\_deterministic57 & <0.001 & <0.001 & <0.001 & <0.001 & <0.001 & <0.001 & <0.001 & <0.001 & <0.001 & <0.001 & <0.001 & <0.001\\
shuffle\_deterministic84 & <0.001 & <0.001 & <0.001 & <0.001 & <0.001 & <0.001 & <0.001 & <0.001 & <0.001 & <0.001 & <0.001 & <0.001\\
shuffle\_even\_odd & <0.001 & <0.001 & <0.001 & <0.001 & <0.001 & <0.001 & <0.001 & <0.001 & <0.001 & <0.001 & <0.001 & <0.001\\
\addlinespace
shuffle\_local10 & <0.001 & <0.001 & <0.001 & <0.001 & <0.001 & <0.001 & <0.001 & <0.001 & <0.001 & <0.001 & <0.001 & <0.001\\
shuffle\_local2 & <0.001 & <0.001 & <0.001 & <0.001 & <0.001 & <0.001 & <0.001 & 0.0036 & <0.001 & 0.002 & 0.0107 & 0.0403\\
shuffle\_local3 & <0.001 & <0.001 & <0.001 & <0.001 & <0.001 & <0.001 & <0.001 & <0.001 & <0.001 & <0.001 & <0.001 & <0.001\\
shuffle\_local5 & <0.001 & <0.001 & <0.001 & <0.001 & <0.001 & <0.001 & <0.001 & <0.001 & <0.001 & <0.001 & <0.001 & <0.001\\
shuffle\_nondeterministic & <0.001 & <0.001 & <0.001 & <0.001 & <0.001 & <0.001 & <0.001 & <0.001 & <0.001 & <0.001 & <0.001 & <0.001\\
\bottomrule
\end{tabular}
\caption{Welch’s t-test comparing each perturbation with shuffle\_control across 12 checkpoints for \textbf{Turkish}, with Bonferroni adjustment.}
\end{table*}

\end{document}